\documentclass[10pt,journal,compsoc]{IEEEtran}
\pdfoutput=1
\ifCLASSOPTIONcompsoc
  \usepackage[nocompress]{cite}
\else
  \usepackage{cite}
\fi
\ifCLASSINFOpdf
\else
\fi
\usepackage{latexsym}
\usepackage{amssymb}
\usepackage{bm}
\usepackage{makecell}
\usepackage{xcolor}
\usepackage{microtype}

\usepackage{amsmath}
\usepackage{subfigure}
\usepackage{graphicx}
\usepackage{booktabs}
\usepackage{multirow}
\usepackage{enumerate}
\usepackage{amsfonts}

\usepackage{url}

\usepackage[ruled,linesnumbered]{algorithm2e}
\newcommand\RoBERTaBASE{RoBERTa$_{\small \textsc{Base}}$}
\newcommand\BertBASE{BERT$_{\small \textsc{Base}}$}

\begin{document}
%
\title{CSS-LM: A Contrastive Framework for Semi-supervised Fine-tuning of Pre-trained Language Models}
%
%
%
%

\author{Yusheng Su, Xu Han, Yankai Lin, Zhengyan Zhang, Zhiyuan Liu, Peng Li, Jie Zhou, Maosong Sun
\IEEEcompsocitemizethanks{\IEEEcompsocthanksitem Yusheng Su, Xu Han, Zhengyan Zhang, Zhiyuan Liu, Maosong Sun are with the Department of Computer Science and Technology, Tsinghua University, Beijing 100084, China.\protect\\
E-mail: \{sys19, hanxu17, zy-z19\}@mails.tsinghua.edu.cn, \protect\\ \{liuzy, sms\}@tsinghua.edu.cn
\IEEEcompsocthanksitem Yankai Lin, Peng Li and Jie Zhou are with the Pattern Recognition Center, WeChat AI Department, Tencent, Beijing 100080, China.\protect\\Email: \{yankailin, patrickpli, withtomzhou\}@tencent.com
\IEEEcompsocthanksitem Zhiyuan Liu is the corresponding author.}}

%
%

\markboth{Journal of \LaTeX\ Class Files,~Vol.~14, No.~8, August~2015}%
{Shell \MakeLowercase{\textit{et al.}}: Bare Demo of IEEEtran.cls for Computer Society Journals}
%



\IEEEtitleabstractindextext{%
\begin{abstract}
Fine-tuning pre-trained language models (PLMs) has demonstrated its effectiveness on various downstream NLP tasks recently. However, in many scenarios with limited supervised data, the conventional fine-tuning strategies cannot sufficiently capture the important semantic features for downstream tasks. To address this issue, we introduce a novel framework (named ``CSS-LM'') to improve the fine-tuning phase of PLMs via contrastive semi-supervised learning. Specifically, given a specific task, we retrieve positive and negative instances from large-scale unlabeled corpora according to their domain-level and class-level semantic relatedness to the task. We then perform contrastive semi-supervised learning on both the retrieved unlabeled instances and original labeled instances to help PLMs capture crucial task-related semantic features. The experimental results show that CSS-LM achieves better results than the conventional fine-tuning strategy on a series of downstream tasks with few-shot settings by up to 7.8\%, and outperforms the latest supervised contrastive fine-tuning strategy by up to 7.1\%. Our datasets and source code will be available to provide more details.
\end{abstract}

}

\maketitle

\IEEEdisplaynontitleabstractindextext

%
\IEEEpeerreviewmaketitle

\IEEEraisesectionheading{\section{Introduction}\label{sec:introduction}}
\IEEEPARstart{P}{RE-TRAINED} language models (PLMs) like BERT \cite{devlin2018bert} and RoBERTa~\cite{liu2019roberta} can learn general language understanding abilities from large-scale unlabeled corpora, and provide informative contextual representations for downstream tasks. In recent years, instead of learning task-oriented models from scratch, it has gradually become a consensus to fine-tune PLMs for specific tasks, which has been demonstrated on various downstream NLP tasks, including dialogue~\cite{zhang2019dialogpt}, summarization~\cite{zhang2019pegasus,liu-lapata-2019-text}, question answering~\cite{joshi2019spanbert,adiwardana2020humanlike}, and relation extraction~\cite{baldini-soares-etal-2019-matching,peng2020learning}.

\begin{figure*}[t]
\centering
\includegraphics[width=1\textwidth]{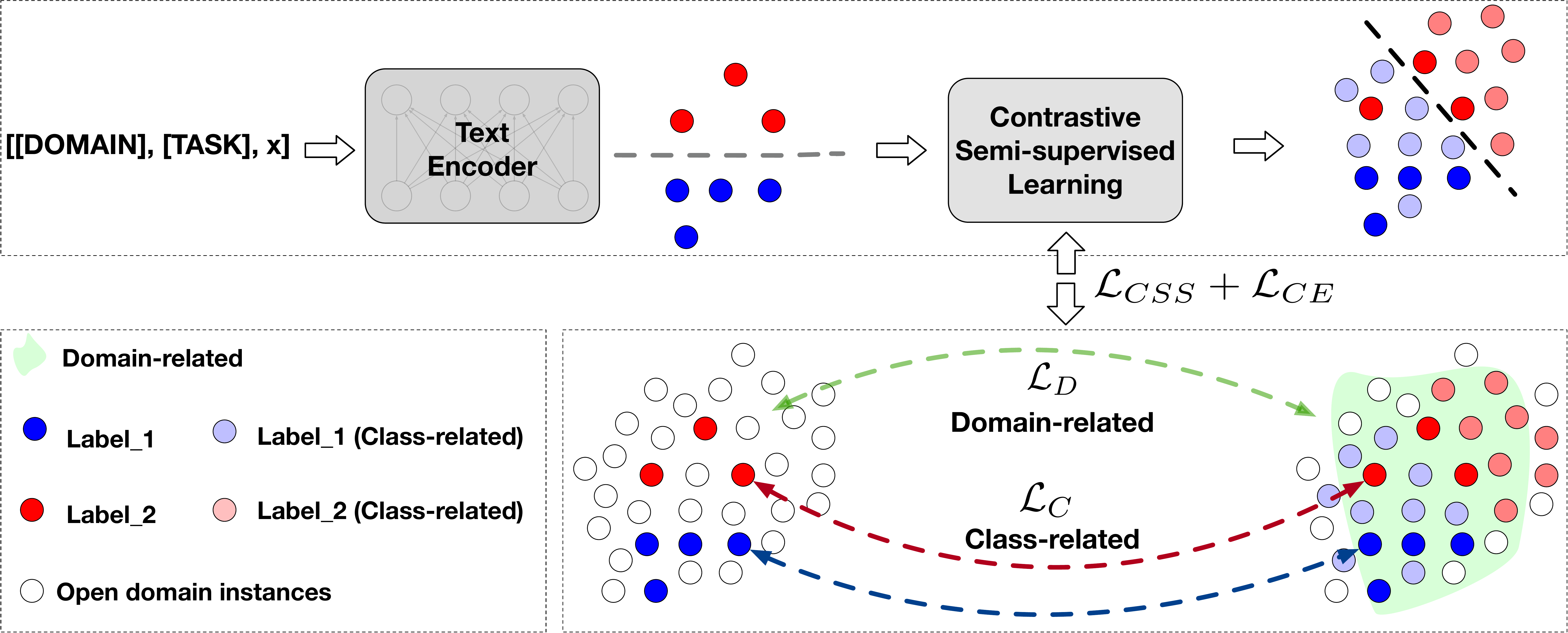}
\caption{The overall framework of CSS-LM, which illustrates how we leverage specific task instances~(red and blue dots) to retrieve task-related instances~(light red and light blue dots in the green cluster) by measuring domain-related and class-related semantics. In the upper figure, by performing contrastive semi-supervised leaning fine-tuning, we can obtain a better decision boundary for the task.}
\label{fig:method}
\end{figure*}

Although fine-tuning PLMs has become a dominant paradigm in the NLP community, it still requires large amounts of supervised data to capture critical semantic features for downstream tasks~\cite{Su_2020,jiang-etal-2020-smart}. Without sufficient supervised data for downstream tasks, the conventional fine-tuning strategy might capture biased features for the downstream tasks that may cause errors or have decision boundary bias in Fig.~\ref{fig:method}. Therefore, fine-tuning PLMs is still challenging in those scenarios with limited data, and cannot be well generalized to many real-world applications whose labeled data is hard and expensive to obtain. Hence, a natural question to ask is: \emph{\textbf{How can we effectively capture crucial semantic features for downstream tasks with limited supervised data?}}

To address this issue, some preliminary works have made some attempts to utilize semi-supervised methods with unlabeled data for fine-tuning PLMs~\cite{du2020selftraining}. Nevertheless, these methods require extra efforts on high-quality labeling to start the semi-supervised learning process. Another way, which can capture crucial semantic features from the limited supervised data in the fine-tuning stage without labeling, is applying contrastive learning~\cite{gunel2021supervised}, by forcing the positive instances to be close to each other in the semantic space, and meanwhile forcing the negative instances to be far away from the positive ones. However, existing contrastive learning methods for enhancing PLMs still lack leveraging the rich large-scale open-domain corpora.

In order to make the use of large-scale open-domain corpora to capture crucial semantic features better, we introduce a novel \textbf{C}ontrastive framework for \textbf{S}emi-\textbf{S}upervised Fine-tuning P\textbf{LM}s (named ``CSS-LM''), which extends the conventional supervised fine-tuning strategy into a semi-supervised form enabling PLMs to leverage extra task-related data from large-scale open-domain corpora without annotating new labels. More specifically, CSS-LM separates the unlabeled data into the positive instances and the negative ones according to these instances' semantic relatedness to downstream tasks. Afterward, CSS-LM applies contrastive learning to let PLMs distinguish the nuances between these instances, so that PLMs can learn informative semantic features not expressed by the limited supervised data of the downstream task. We give an intuitive motivation description in Fig.~\ref{fig:method} to show how our framework could better capture crucial semantics from unsupervised data to make final fine-tuned pre-trained language models more discriminative and have better decision boundaries. 

As to use all unlabeled data is nearly impossible, we require to find the most informative positive and negative instances for our contrastive learning. Therefore, one key challenge of CSS-LM is how to measure the semantic relatedness of unlabeled instances to downstream tasks. 

In this paper, we consider both domain-level and class-level semantic relatedness of instances to downstream tasks. More specifically, we encode unlabeled instances into the domain-level and class-level semantic spaces, respectively, and define the domain-level and class-level relatedness as the similarity of their representations in the corresponding semantic spaces. When performing contrastive semi-supervised learning for the domain-level representations, all supervised instances of downstream tasks are used as the initialized positive instances. After that, we continue to retrieve unlabeled instances closest and farthest from the existing positive instances as new positive and negative instances. When performing contrastive semi-supervised learning for the class-level representations, we apply operations similar to learning domain-level representations; the only difference is that learning class-level representations requires considering fine-grained class-level semantics rather than coarse-grained domain-level semantics when selecting positive and negative instances.


In the experiments, the results on three typical classification tasks, including sentiment classification, intent classification, and relation classification, show that our proposed framework can outperform the conventional fine-tuning strategy, the conventional semi-supervised strategies, and the latest supervised contrastive fine-tuning strategies under the limited supervised data settings. These results explicitly indicate a promising direction of utilizing unlabeled data for fine-tuning PLMs based on contrastive semi-supervised learning.

To summarize, our major contributions are as follows:
\begin{enumerate}[(1)]
\item We propose a contrastive semi-supervised framework CSS-LM, which can better leverage unlabeled instances from open-domain corpora to capture task-related features and enhance the model performance on the downstream tasks.
\item CSS-LM is free to the domain dependence, which can efficiently capture domain information from open-domain corpora and retrieve domain-related instances.
\item We conduct experiments on six classification datasets. The experimental results show that CSS-LM outperforms various typical fine-tuning models under the few-shot settings. Besides, sufficient empirical analyses of our retrieval mechanism and retrieval instances demonstrate that contrastive semi-supervised learning is more helpful than semi-supervised learning (pseudo labeling) to learn from unlabeled data. 
\end{enumerate}

\section{Related Work}
\label{sec:related_work}
\subsection{Pre-trained Language Models}
\label{secsec:pretrained_language_models}

Various recent PLMs like BERT~\cite{devlin2018bert}, RoBERTa~\cite{liu2019roberta} and XLNet~\cite{yang2019xlnet}, provide a new perspective for NLP models to utilize a large amount of open-domain unlabeled data. Inspired by these works, a series of works have designed specific self-supervised learning objectives to help PLMs learn specific abilities in the pre-training phase, such as representing token spans~\cite{joshi2019spanbert} and entities~\cite{zhang2019ernie,su2020cokebert,Peters2019KnowledgeEC,sun2019ernie,Peters2019KnowledgeEC}, conferential reasoning~\cite{ye2020corefbert}, multi-lingualism~\cite{lample2019cross}, multi-modality~\cite{li2019visualbert,Li_2020,su2020vlbert,lu2019vilbert,Tan_2019}, etc. Besides, some PLMs~\cite{Beltagy2019SciBERTAP,huang2020clinicalbert,10.1093/bioinformatics/btz682} are devoted to learning specific domain semantics by pre-training on the specific domain corpora, and these works demonstrate their effectiveness as well.

As PLMs are aimless concerning various downstream tasks in the pre-training stage. Hence, to adapt PLMs for a specific task requires a fine-tuning on extra supervised data of the tasks. Specifically, fine-tuning often replaces the top layers of PLMs with a specific task sub-network, and continues to update the parameters with the supervised data. Fine-tuning PLMs has also demonstrated its effectiveness on various downstream tasks, including dialogue~\cite{zhang2019dialogpt}, summarization~\cite{zhang2019pegasus,liu-lapata-2019-text}, question answering~\cite{adiwardana2020humanlike}, and relation extraction~\cite{baldini-soares-etal-2019-matching,peng2020learning}. 

However, without sufficient supervised data, the conventional fine-tuning methods cannot effectively capture useful features for downstream tasks, which may lead to the side effect on performance. To address this issue, a series of works focus on exploring various heuristics during tuning the parameters of PLMs. Howard and Ruder et al.~\cite{howard-ruder-2018-universal} gradually unfreeze the layers of PLMs with a heuristic learning rate schedule to enhance the fine-tuning performance of PLMs. Then, Peters et al.~\cite{peters-etal-2019-tune} study which layers of PLMs should be adapt or freeze during the fine-tuning stage. Houlsby et al.~\cite{houlsby2019parameter} and Stickland et al.~\cite{Stickland019} leverage some additional layers to PLMs and update parameters of specific additional layers during the fine-tuning phase. 

Besides, some preliminary works have made some attempts to utilize unlabeled data for fine-tuning PLMs: Gu et al.~\cite{gu2020train} conduct selective language modeling with unlabeled data to focus on the semantic features related to the fine-tuning tasks. Gururangan et al.~\cite{Gururangan_2020} propose a framework to retrieve task-related data from large-scale in-domain corpora to enhance fine-tuning PLMs. The in-domain instances often meet an individual margin probability distribution over instances, and they thus have correlated semantics in the feature space that is beneficial for specific tasks. Du et al.~\cite{du2020selftraining} further introduce a text retriever trained on a large amount of supervised data to retrieve task-specific in-domain data from large-scale open-domain corpora. 

Existing works for enhancing fine-tuning rely on pre-defined in-domain corpora, massive supervised data, or extra efforts on labeling, limiting them to be applied to broad real-world applications. Unlike these works, our proposed contrastive semi-supervised framework can automatically and iteratively utilize large-scale open-domain data to improve fine-tuning under the few-shot settings without annotating extra data.

\subsection{Contrastive Learning}
\label{secsec:contrastive_learning}
Unlike conventional discriminative methods that learn a mapping to labels and generative methods that reconstruct input instances, contrastive learning is a learning paradigm based on comparing. Specifically, contrastive learning can be considered as learning by comparing among different instances instead of learning from individual instances one at a time. The comparison can be performed between positive pairs of "similar" instances and negative pairs of "dissimilar" instances. The early efforts for self-supervised contrastive learning have led to significant advances in NLP~\cite{logeswaran2018efficient,kong2019mutual,zhang2020unsupervised,wu2020clear,peng2020learning} and CV~\cite{chen2020simple,He_2020,Tian_2020,Kolesnikov_2019_CVPR,NIPS2013_db2b4182,JMLR:v13:gutmann12a} tasks. Nevertheless, self-supervised contrastive learning still has a limitation: it cannot utilize the supervised data of downstream tasks and sufficiently capture the fine-grained semantics to specific classes. Intuitively, self-supervised contrastive learning tends to distinguish instances rather than to classify instances. 

Therefore, Khosla et al.~\cite{khosla2020supervised} propose supervised contrastive learning to leverage the supervised instances of downstream tasks. More recently, Gunel et al.~\cite{gunel2021supervised} verify the effectiveness of supervised contrastive learning in the fine-tuning stage of PLMs. However, it only considers the limited amounts of the supervised data, and ignores the potential information distributed in the unlabeled data. 

In this paper, our approach is a general semi-supervised fine-tuning framework based on contrastive learning, which could capture richer semantics from unlabeled data and align features with the supervised data of downstream tasks, simultaneously distinguishing and classifying instances.

\section{Methodology}
\label{sec:methodology}
In this work, we mainly focus on fine-tuning PLMs for classification tasks. Unlike conventional fine-tuning methods aiming to learn features by classifying the limited labeled training data, CSS-LM leverages unlabeled data in open-domain corpora to capture better features. More specifically, as illustrated in Fig.~\ref{fig:method}, CSS-LM utilizes contrastive semi-supervised learning in the fine-tuning phase of PLMs, aiming to identify all crucial semantic features to distinguish the instances of different domains and classes with both labeled and unlabeled data, and further obtain the better decision boundary. 

CSS-LM consists of five important modules, including (1)~contrastive semi-supervised learning, (2)~informative instance retrieval, (3)~semantic representation learning, (4)~downstream task fine-tuning, and (5)~efficient representation updating. In this section, we will first give some essential notations and then introduce these essential modules. 

\subsection{Notations}
\label{ssec:notations}

Given a specific downstream classification task, we denote its class set as $\mathcal{Y}$, and its training set as $\mathcal{T} = \{(x^1_{\mathcal{T}}, y^1_{\mathcal{T}}), (x^2_{\mathcal{T}}, y^2_{\mathcal{T}}), \ldots, (x^N_{\mathcal{T}}, y^N_{\mathcal{T}})\}$, where $y^i_{\mathcal{T}} \in \mathcal{Y}$ is the supervised annotation of the instance $x^i_{\mathcal{T}}$ and $N$ is the instance number of the training set. Under the few-shot setting, for each class $y \in \mathcal{Y}$, we denote the number of instances whose class is $y$ as $K_y$. 

As our framework will utilize large-scale unsupervised open-domain corpora, we denote the open-domain unlabeled corpora as $\mathcal{O}=\{x_{\mathcal{O}}^{1}, x_{\mathcal{O}}^{2}, \cdots, x_{\mathcal{O}}^{M}\}$, where $M$ is the instance number of unlabeled corpora. We use the bold face to indicate the representation of an instance computed by PLMs, e.g., the representation of $x^i_{\mathcal{T}}$ is $\mathbf{x}^i_{\mathcal{T}}$. As $\mathbf{x}^i_{\mathcal{T}}$ is computed by the PLM-based encoder, we denote the encoder as $\texttt{Enc}(\cdot)$, which will be introduced in~\ref{sssec:pre-trained_language_model_encoders}. 

Note that our framework is applicable to any PLMs. In experiments, we select BERT~\cite{devlin2018bert} and RoBERTa~\cite{liu2019roberta} as our encoders, considering they are the state-of-the-art and widely-used ones of existing PLMs.



\subsection{Contrastive Semi-supervised Learning}
\label{ssec:contrastive_semisupervised_learning}

Given the training set of a specific downstream task $\mathcal{T}$, we first introduce how to apply contrastive learning for supervised fine-tuning:
\begin{equation}
\begin{aligned}
\label{equ:sc}
    \mathcal{L}_{CS} &= -\sum_{i=1}^N \frac{\sum_{j\in \mathcal{C}(i)} f(\mathbf{x}^i_{\mathcal{T}},\mathbf{x}^j_{\mathcal{T}}) - \log Z^i}{|\mathcal{C}(i)|},\\
    Z^i &= \sum_{k=1}^N e^{ f(\mathbf{x}^i_{\mathcal{T}},\mathbf{x}^k_{\mathcal{T}})},
\end{aligned}
\end{equation}
where $\mathcal{C}(i) = \{j|j\neq i, y^j_{\mathcal{T}}=y^i_{\mathcal{T}}\}$ is the index set of instances that share the same label with the instance $x^i_{\mathcal{T}}$, and $f(\cdot, \cdot)$ is the similarity function. Considering many downstream tasks only have limited supervised data, we apply contrastive semi-supervised learning in the fine-tuning phase of PLMs instead of Eq.~(\ref{equ:sc}). Specifically, we retrieve positive and negative instances from the open-domain corpora $\mathcal{O}$ for each supervised instance. 

We denote the positive instance index set of $x^i_{\mathcal{T}}$ as $\mathcal{P}(i)$ and the negative one as $\mathcal{N}(i)$. Formally, the contrastive semi-supervised objective for fine-tuning PLMs is:
\begin{equation}
\label{eq:css}
\begin{aligned}
    &\mathcal{L}_{CSS} = - \sum_{i=1}^N  \big( \frac{\sum_{j\in \mathcal{C}(i)}
 f(\mathbf{x}^i_{\mathcal{T}},\mathbf{x}^j_{\mathcal{T}}) - \log Z_s^i}{|\mathcal{C}(i)|}  + \\
 &\frac{\sum_{j\in \mathcal{P}(i)}
 f(\mathbf{x}^i_{\mathcal{T}},\mathbf{x}^j_{\mathcal{O}}) -\log Z_s^i}{|\mathcal{P}(i)|} +\\
  & \frac{\sum_{j\in \mathcal{P}(i)}\sum_{k\in \mathcal{P}(i),j\neq k}  f(\mathbf{x}^j_{\mathcal{O}},\mathbf{x}^k_{\mathcal{O}}) - \log Z_u^j}{(|\mathcal{C}(i)|+1)\times|\mathcal{P}(i)|\times(|\mathcal{P}(i)|-1)}  \big),
\end{aligned}
\end{equation}
where $Z_s^i$ and $Z_u^j$ are calculated as:
\begin{equation}
\begin{aligned}
    Z_s^i=& \sum_{k=1}^{N}e^{ f(\mathbf{x}^i_{\mathcal{T}},\mathbf{x}^k_{\mathcal{T}})} +  \sum_{k \in \mathcal{N}(i) \cup \mathcal{P}(i)}e^{ f(\mathbf{x}^i_{\mathcal{T}},\mathbf{x}^k_{\mathcal{O}})},\\
    Z_u^j=& \sum_{k \in \mathcal{P}(i), j\neq k}e^{ f(\mathbf{x}^j_{\mathcal{O}},\mathbf{x}^k_{\mathcal{O}})}+ \sum_{k \in \mathcal{N}(i)}e^{ f(\mathbf{x}^j_{\mathcal{O}},\mathbf{x}^k_{\mathcal{O}})}.
\label{eqn:semi-supervised-contrastive-learning}
\end{aligned}
\end{equation}
Intuitively, by applying contrastive semi-supervised learning with Eq.~(\ref{eq:css}), we can simultaneously consider the similarities among both supervised and retrieved instances, which can let PLMs well capture semantics from the unlabeled data of the task $\mathcal{T}$.

\subsection{Informative Instance Retrieval}
\label{ssec:informative_instance_retrieval}

Given a supervised instance $x^i_{\mathcal{T}}$, we retrieve the most informative instances from the open-domain corpora $\mathcal{O}$ to build the positive instance index set $\mathcal{P}(i)$ and the negative one $\mathcal{N}(i)$. 

A straightforward retrieval solution is to regard the most similar instances as positive instances and the most dissimilar ones as negative instances, i.e., retrieve the instances $x^*_{\mathcal{O}}$ according to $f(\mathbf{x}^i_{\mathcal{T}},\mathbf{x}^*_{\mathcal{O}})$. However, this straightforward solution is too coarse to select those most informative instances, since the instances of the open-domain corpora may belong to various quite different domains and classes. We cannot know which semantic levels would make the greater contribution with the coarse-grained function $f(\mathbf{x}^i_{\mathcal{T}},\mathbf{x}^*_{\mathcal{O}})$ measuring. 

Thus, instead of using $f(\cdot)$ to retrieve instances, we introduce a similarity function $f_T(\cdot)$ for retrieving the most informative instances by empirically considering the semantic relatedness between instances from two perspectives: domain-level and class-level similarities. More specifically, given instances $x^i_{\mathcal{T}},x^*_{\mathcal{O}}$, $f_T(\cdot)$ obtains the instance relatedness by calculating the summation of the domain relatedness and class relatedness as:
\begin{equation}
\begin{aligned}
\label{equ:retrieveal}
    f_{T}(\mathbf{x}^i_{\mathcal{T}},\mathbf{x}^*_{\mathcal{O}}) = f_{D}(\mathbf{d}^i_{\mathcal{T}},\mathbf{d}^*_{\mathcal{O}})+f_{C}(\mathbf{c}^i_{\mathcal{T}},\mathbf{c}^*_{\mathcal{O}}),
\end{aligned}
\end{equation}
where $f_D(\cdot)$ is the domain similarity function, $f_C(\cdot)$ is the class similarity function, $\mathbf{d}^i_{\mathcal{T}},\mathbf{d}^*_{\mathcal{T}}$ are the domain-level representations, and $\mathbf{c}^i_{\mathcal{T}},\mathbf{c}^*_{\mathcal{T}}$ are the class-level representations. In this way, we could build the positive instance index set $\mathcal{P}(i)$ and the negative one $\mathcal{N}(i)$. 

However, how to obtain domain-level and class-level representations is still a problem. Next, we will introduce how to encode instances into domain-level and class-level spaces, and how to learn their representations respectively under our contrastive semi-supervised framework.

\subsection{Semantic Representation Learning}
\label{ssec:semantic_representation_learning}
This section gives the details of the text encoder and introduces how to learn domain-level and class-level instance representations.

\begin{figure}[t]
\centering
\includegraphics[width=0.5\textwidth]{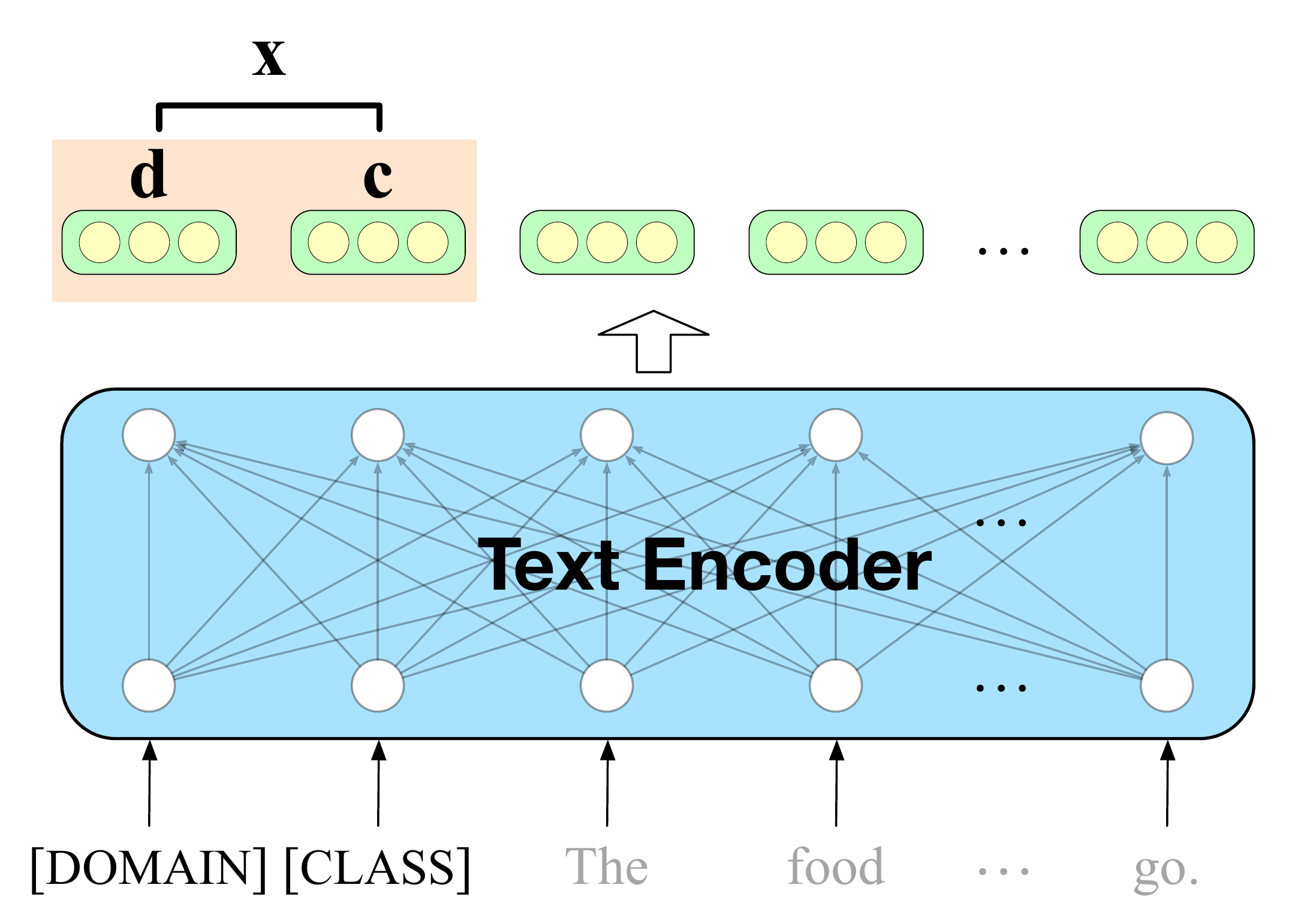}
\caption{Given an instance, we add two special tokens \texttt{[DOMAIN]} and \texttt{[CLASS]} into the sequence, and then input the sequence into the PLM-based encoder. $\mathbf{d}$ and $\mathbf{c}$ are the representations of \texttt{[DOMAIN]} and \texttt{[CLASS]} respectively. $\mathbf{x}$ is the concatenation of the two special token representations.}
\label{fig:plm}
\end{figure}

\subsubsection{Encoder Based on Pre-trained Language Models}
\label{sssec:pre-trained_language_model_encoders}
Given an instance $x$ (either a supervised instance or a unlabeled one), as show in Fig.~\ref{fig:plm}, we first add two special tokens in front of the input sequence of $x$, i.e., $\bar{x} =\big[[\texttt{DOMAIN}], [\texttt{CLASS}], x \big]$. Then, we input $\bar{x}$ into the encoder, which is a multi-layer bidirectional Transformer encoder, such as \BertBASE~\cite{devlin2018bert} and \RoBERTaBASE~\cite{liu2019roberta}, and use the output representations of $[\texttt{DOMAIN}]$ and $[\texttt{CLASS}]$ as the domain-level representation $\mathbf{d}$ and the class-level representation $\mathbf{c}$ respectively. We then define the whole instance representation $\mathbf{x}$ as the concatenation of the domain-level and class-level representations as: 
\begin{equation}
\label{eqn:text_encoder}
    \mathbf{x} = [\mathbf{d};\mathbf{c}] = \texttt{Enc}(\bar{x}),
\end{equation}
where $\mathbf{d}$ and $\mathbf{c}$ are the representations of $\texttt{[DOMAIN]}$ and $\texttt{[CLASS]}$ respectively, $[\cdot;\cdot]$ is the concatenation of representations, and $\texttt{Enc}(\cdot)$ indicates the PLM-based text encoder. 


\subsubsection{Domain-level Representations} 
\label{sssec:domain-level_representations}
We hope that domain-level representations can push instances with different domains far away, and meanwhile, cluster those instances with similar domains. Hence, the whole learning objective is:
\begin{equation}
\begin{aligned}
      \mathcal{L}_{D} &= - \sum_{i=1}^N \big( \frac{\sum_{j=1}^N
 f_{D}(\mathbf{d}_{\mathcal{T}}^i, \mathbf{d}_{\mathcal{T}}^j) - \log Z_d^i}{N}\\
  & + \frac{\sum_{j\in \mathcal{P}_D(i)}f_{D}(\mathbf{d}_{\mathcal{T}}^i,\mathbf{d}^j_{\mathcal{O}}) - \log Z_d^i}{|\mathcal{P}_D(i)|} \big),\\
  Z_d^i &= \sum_{k=1}^{N} e^{ f_{D}(\mathbf{d}_{\mathcal{T}}^i,\mathbf{d}_{\mathcal{T}}^k)}
  + \sum_{k=1}^Me^{f_{D}(\mathbf{d}_{\mathcal{T}}^i,\mathbf{d}_{\mathcal{O}}^k)},
\end{aligned}
\end{equation}
where $\mathcal{P}_{D}(i)$ is the domain-related positive instance index set of $x_{\mathcal{T}}^i$. $x_{\mathcal{T}}^i$ is retrieved from the open-domain corpora according to the similarity between domain-level representations computed by $f_{D}(\cdot)$. Besides the instances mentioned in $x_{\mathcal{T}}^i$, we regard all other instances of the open-domain corpora $\mathcal{O}$ as domain-related negative instances.

\subsubsection{Class-level Representations}
\label{sssec:class-level_representations}
Class-level representations aim to distinguish the semantic difference between the different classes of the task, and push away those instances not related to one specific class. Hence, the learning objective is formulated as:
\begin{equation}
\begin{aligned}
      \mathcal{L}_{C}&= - \sum_{i=1}^N \big( \frac{\sum_{j=1}^N
 f_{C}(\mathbf{c}_{\mathcal{T}}^i, \mathbf{c}_{\mathcal{T}}^j) - \log Z_c^i}{N}\\
  & + \frac{\sum_{j\in \mathcal{P}_C(i)}f_{C}(\mathbf{c}_{\mathcal{T}}^i,\mathbf{c}^j_{\mathcal{O}}) - \log Z_c^i}{|\mathcal{P}_C(i)|} \big),\\
  Z_c^i &= \sum_{k=1}^{N} e^{ f_{C}(\mathbf{c}_{\mathcal{T}}^i,\mathbf{c}_{\mathcal{T}}^k) }
  + \sum_{k\in \mathcal{P}_C(i) \cup \mathcal{N}_C(i)} e^{f_{C}(\mathbf{d}_{\mathcal{T}}^i, \mathbf{d}_{\mathcal{O}}^k)},
\end{aligned}
\end{equation}
where $\mathcal{P}_C(i)$ is the class-related positive instance index set of $x^i_{\mathcal{T}}$ retrieved from the open-domain corpora according to the similarity between the class-level representations computed by $f_C(\cdot)$. Note that, given the class of $x^i_{\mathcal{T}}$, we take the supervised instances belonging to all other classes and the retrieved positive instances of all other classes as the negative instance set of $x^i_{\mathcal{T}}$. The index set of these negative instances of $x^i_{\mathcal{T}}$ is denoted as $\mathcal{N}_C(i)$.

\subsection{Efficient Representation Updating}
\label{sec:efficient_representation_updating}
Note that we set $\mathcal{P}(i)$, $\mathcal{P}_D(i)$ and $\mathcal{P}_C(i)$ as empty sets at the beginning of the fine-tuning phase, since at that time, three kinds of representations are not well trained. Then, we will iteratively retrieve instances from the open-domain corpora to expand these sets. Since we found the PLM parameters will not change vastly during fine-tuning for the downstream task, for computational efficiency, we only update the parts of instance representations in the open-domain corpora $\mathcal{O}$, which are retrieved in every step, rather than updating the whole instance representations.

\subsection{Downstream Task Fine-tuning and Optimization}
\label{ssec:downstream_task_finetuning_and_optimization}

In classification tasks, we are devoted to finding effective decision boundaries with semantic representations. Given the training instances $(x_\mathcal{T}^{i},y^{i}_{\mathcal{T}}) \in \mathcal{T}$, we encode $x_\mathcal{T}^{i}$ to $\textbf{x}_\mathcal{T}^{i}$ with the text encoder $\texttt{Enc}(\cdot)$. Similar to the conventional PLM fine-tuning, we apply the cross-entropy loss to learn task-specified classifier as:
\begin{equation}
\label{eq:loss}
\begin{aligned}
    \mathcal{L}_{CE} = -\sum_{i=1}^{N}  
    \log f_{task}(\textbf{x}_\mathcal{T}^{i}, y_{\mathcal{T}}^{i}),
\end{aligned}
\end{equation}
where $f_{task}(\cdot)$ is the neural layers for specific tasks built on the PLMs to compute the probability $p(y_{\mathcal{T}}^{i}|x_{\mathcal{T}}^{i})$.

\begin{algorithm}[]
\SetAlgoLined
\LinesNumberedHidden
\KwData{Training set $\mathcal{T}$; open-domain corpora $\mathcal{O}$; development set $\mathcal{E}$;}
\KwResult{The optimal CSS-LM parameters $\theta$}
\textbf{Initialization}: Parameters of CSS-LM $\theta$ [Refer to ~\ref{ssec:experimental_settings}]; task-related sets $\mathcal{P}(i)$, $\mathcal{N}(i)$; domain-related set $\mathcal{P}_{D}(i)$; class-related sets $\mathcal{P}_C(i)$, $\mathcal{N}_C(i)$\; 
\For{$epoch \gets [0,..,E]$}{
\While{$i \in N$}{ 
        Encode an instance of the training set:
        $\mathbf{x}^i_{\mathcal{T}} = [\mathbf{d}^{i}_\mathcal{T};\mathbf{c}^{i}_\mathcal{T}] = \texttt{Enc}(\bar{x}^i_{\mathcal{T}})$~[Refer to~\ref{sssec:pre-trained_language_model_encoders}]
        \; \
        
        Retrieve the domain-related instances by $f_{D}(\cdot)$ \\ and update $\mathcal{P}_{D}(i),\mathcal{N}_{D}(i)$, \\
        $[\mathbf{d}^{j}_\mathcal{O};\mathbf{c}^{j}_\mathcal{O}] = \texttt{Enc}(\bar{x}^j_{\mathcal{O}})$, where\\ $\bar{x}^j_{\mathcal{O}} \in\{\mathcal{P}_{D}(i),\mathcal{N}_{D}(i)\}$, \\
        $\mathbf{d}^{j}_\mathcal{O} \gets Select_{\textbf{d}}([\mathbf{d}^{j}_\mathcal{O};\mathbf{c}^{j}_\mathcal{O}])$, \\
        $\mathbf{d}^{i}_\mathcal{T} \gets Select_{\textbf{d}}([\mathbf{d}^{i}_\mathcal{T};\mathbf{c}^{i}_\mathcal{T}])$, \\
        Perform $\mathcal{L_{D}}$~[Refer to~\ref{sssec:domain-level_representations}]
        \; \
        
        Retrieve the class-related instances by $f_{C}(\cdot)$ \\ and update $\mathcal{P}_{C}(i),\mathcal{N}_{C}(i)$, \\ 
        $[\mathbf{d}^{j}_\mathcal{O};\mathbf{c}^{j}_\mathcal{O}] = \texttt{Enc}(\bar{x}^j_{\mathcal{O}})$, where\\ $\bar{x}^j_{\mathcal{O}} \in\{\mathcal{P}_{C}(i),\mathcal{N}_{C}(i)\}$, \\
        $\mathbf{c}^{j}_\mathcal{O} \gets Select_{\textbf{c}}([\mathbf{d}^{j}_\mathcal{O};\mathbf{c}^{j}_\mathcal{O}])$, \\
        $\mathbf{c}^{i}_\mathcal{T} \gets Select_{\textbf{c}}([\mathbf{d}^{i}_\mathcal{T};\mathbf{c}^{i}_\mathcal{T}])$, \\
        Perform $\mathcal{L_{C}}$~[Refer to~\ref{sssec:class-level_representations}]
        \; \
        
        Retrieve the task-related instances by $f_{T}(\cdot)$ \\
        and update $\mathcal{P}(i),\mathcal{N}(i)$ [Refer to~\ref{ssec:informative_instance_retrieval}], \\
        $ \mathbf{x}^{j}_\mathcal{O} = \texttt{Enc}(\bar{x}^{j}_{\mathcal{O}})$, where $\bar{x}^{j}_{\mathcal{O}}$ $\in\{\mathcal{P}(i),\mathcal{N}(i)\}$, \\
        $\mathbf{x}^{j}_\mathcal{O} \gets \mathbf{x}^{j}_\mathcal{O}$, \\
        $\mathbf{x}^{i}_\mathcal{T} \gets \mathbf{x}^{i}_\mathcal{T}$, \\
        Perform $\mathcal{L_{CSS}}$~[Refer to~\ref{ssec:contrastive_semisupervised_learning}]
        \; \
        
        Leverage $\texttt{Enc}(\cdot)$ to update representations of \\
        $\small{
        \bar{x}^{j}_{\mathcal{O}}\in\{\mathcal{P}_{D}(i), \mathcal{N}_{D}(i), \mathcal{P}_C(i),\mathcal{N}_C(i),\mathcal{P}(i),\mathcal{N}(i)\} 
        }$ \\
        to the open-domain corpora representation \\ 
        set~[Refer to~\ref{sec:efficient_representation_updating}]
        \; \
        
        Calculate the total loss: $\mathcal{L} = \mathcal{L}_{CSS} + \mathcal{L}_D + \mathcal{L}_C + \mathcal{L}_{CE}$ 
        [Refer to~\ref{ssec:downstream_task_finetuning_and_optimization}]\; \
        
        Compute gradient $\nabla_{\theta}\mathcal{L}(\theta;\mathcal{T})$\;
        Update parameters $\theta = \theta - \lambda\cdot\nabla_{\theta}\mathcal{L}(\theta;\mathcal{T})$\;
    }
    Save $\theta$ as $\theta_{epoch}$ in every epoch\;
}
\textbf{Return}: the optimal $\theta_{epoch}$ in the development set $\mathcal{E}$\;
\caption{Contrastive Semi-supversied Learning}
\label{algorithm:css-lm}
\end{algorithm}

We optimize the domain-level, class-level, and the whole instance representations jointly. Therefore, the overall learning objective is defined as the sum of four losses:
\begin{equation}
    \mathcal{L} = \mathcal{L}_{CSS} + \mathcal{L}_D + \mathcal{L}_C + \mathcal{L}_{CE},
\end{equation}
where $\mathcal{L}_{CSS}$ is the contrastive semi-supervised loss, both $\mathcal{L}_D$ and $\mathcal{L}_C$ are the functions to let the encoder learn domain-level and class-level representations respectively. $ \mathcal{L}_{CE}$ is the conventional fine-tuning cross-entropy loss.

The overall detail can refer to Algorithm~\ref{algorithm:css-lm}. Given an instance in each step, first, CSS-LM retrieves domain-related, class-related, and task-related instances, then delivers to the corresponding positive and negative instance index set. Second, CSS-LM updates the representations of retrieved instances to the open-domain corpora representation set. Finally, CSS-LM leverage these instances to perform contrastive semi-supervised to obtain model parameters in every epoch. We will evaluate the CSS-LM on the development set every epoch and choose the optimal one as our model in downstream tasks.

\begin{table*}[t]
\caption{\label{table:dataset} The details of the datasets used in our experiments. To build few-shot settings, we sample $N= K_{y} \times \mathcal{|Y|}$ instances from the original training set, where $K_{y}$ is the sampled instance number for each class and $\mathcal{|Y|}$ is the number of class types.}
\begin{center} 
\small
\scalebox{0.85}
{
{
\begin{tabular}{ccrrrrrr}
\toprule
    \multicolumn{1}{c}{Task} &  \multicolumn{1}{c}{Domain} & \multicolumn{1}{c}{Dataset} & \multicolumn{1}{c}{$\mathcal{|Y|}$} & \multicolumn{1}{c}{\#Train} & \multicolumn{1}{c}{\#Test} & \multicolumn{1}{c}{\#Dev} & \multicolumn{1}{c}{Class types} \\
    \midrule
    \multirow{2}*{\shortstack{\\Sentiment\\ Classification}} & \multicolumn{1}{c}{Review} & \multicolumn{1}{c}{SemEval} & \multicolumn{1}{c}{3} & \multicolumn{1}{c}{4,665} & \multicolumn{1}{c}{2,426} & \multicolumn{1}{c}{4,665} & \multicolumn{1}{c}{positive, neutral, negative} \\
    \cmidrule{2-8} &
    \multicolumn{1}{c}{Review} & \multicolumn{1}{c}{SST-5} & \multicolumn{1}{c}{5} & \multicolumn{1}{c}{8,544} & \multicolumn{1}{c}{2,210} & \multicolumn{1}{c}{1,101} & \multicolumn{1}{c}{v. pos., positive, neutral, negative, v. neg.} \\
    \midrule
    \multirow{2}*{\shortstack{\\Intent\\Classification}} & \multicolumn{1}{c}{Multi} & \multicolumn{1}{c}{Scicite} & \multicolumn{1}{c}{3} & \multicolumn{1}{c}{7,320} & \multicolumn{1}{c}{1,861} & \multicolumn{1}{c}{916} & \multicolumn{1}{c}{result, method, background}\\
    \cmidrule{2-8} &
    \multicolumn{1}{c}{CS} & \multicolumn{1}{c}{ACL-ARC} & \multicolumn{1}{c}{6} & \multicolumn{1}{c}{1,688} & \multicolumn{1}{c}{139} & \multicolumn{1}{c}{114} & \multicolumn{1}{c}{\shortstack{\\background, uses, motivation, compareOrcontrast, extends, future}}\\
    \midrule
    \multirow{2}*{\shortstack{\\Relation\\Classification}} & 
    \multicolumn{1}{c}{CS} & \multicolumn{1}{c}{SciERC} & \multicolumn{1}{c}{7} & \multicolumn{1}{c}{3,219} & \multicolumn{1}{c}{974} & \multicolumn{1}{c}{455} & 
    \multicolumn{1}{c}{\shortstack{\\part-of, conjunction, hyponymy, used-for, feature-of, compare, evaluate-for}}\\
    \cmidrule{2-8} &
    \multicolumn{1}{c}{BIO} & \multicolumn{1}{c}{ChemProt} & \multicolumn{1}{c}{13} & \multicolumn{1}{c}{4,169} & \multicolumn{1}{c}{3,469} & \multicolumn{1}{c}{2,422} & 
    \multicolumn{1}{c}{\shortstack{\\substrate, antagonist, indirect-upregulator, activator, indirect-downregulator, \\ inhibitor, upregulator, downregulator, product-of, agonist, agonist-activator}}\\
\bottomrule
\end{tabular}}}
\end{center}  
\end{table*}

\section{Experiments}
\label{sec:experiments}

In this section, we would first introduce the datasets, the experimental settings, and the details of the baseline models used in our experiments. After that, we give some empirical analyses to show the effectiveness of our contrastive semi-supervised learning, indicating the promising results of leveraging unlabeled instances. Then, we perform some ablation studies to show which level of semantic relatedness mainly contributes to CSS-LM and the influence of the retrieved size. Finally, we perform visualization and case studies for a more intuitive observation.

\begin{table*}[t]
\caption{\label{table:NLU_task_table} The results~(\%) of various fine-tuning methods on six classification datasets. All fine-tuning strategies are applied on \RoBERTaBASE~and \BertBASE~models and set $K_{y}=16$ for the few-shot experimental settings.}
\begin{center} 
\small
\scalebox{0.9}
{
{
\begin{tabular}{l|cc|cc|cc|cc|cc|cc}
\toprule
  \textbf{Task} & \multicolumn{4}{c|}{\textbf{Sentiment Classification}} & \multicolumn{4}{c|}{\textbf{Intent Classification}} & \multicolumn{4}{c}{\textbf{Relation Classification}}
  \\
  \midrule
  \textbf{Dataset} & 
  \multicolumn{2}{c}{\textbf{SemEval}} & \multicolumn{2}{c|}{\textbf{SST-5}}  & \multicolumn{2}{c}{\textbf{SciCite}} & \multicolumn{2}{c|}{\textbf{ACL-ARC}} & \multicolumn{2}{c}{\textbf{SciERC}} & \multicolumn{2}{c}{\textbf{ChemProt}}
  \\
  \midrule
  \textbf{Base Model} & 
  \multicolumn{1}{c}{\texttt{RoBERTa}} & \multicolumn{1}{c|}{\texttt{BERT}} &
  \multicolumn{1}{c}{\texttt{RoBERTa}} & \multicolumn{1}{c|}{\texttt{BERT}} &
  \multicolumn{1}{c}{\texttt{RoBERTa}} & \multicolumn{1}{c|}{\texttt{BERT}} &
  \multicolumn{1}{c}{\texttt{RoBERTa}} & \multicolumn{1}{c|}{\texttt{BERT}} &
  \multicolumn{1}{c}{\texttt{RoBERTa}} & \multicolumn{1}{c|}{\texttt{BERT}} &
  \multicolumn{1}{c}{\texttt{RoBERTa}} & \multicolumn{1}{c}{\texttt{BERT}}
  \\
  \midrule
\multicolumn{13}{c}{Fine-tune on the whole training set}
\\
\midrule
Standard~\cite{liu2019roberta} & {89.1} & {87.4} & {56.8} & {54.1} & {86.0} & {84.8} & {81.3} & {77.5} & {88.7} & {88.6} & {82.3} & {80.6}\\
SCF~\cite{gunel2021supervised} & {88.9} & {86.9} & {57.4} & {53.6} & \textbf{86.5} & \textbf{85.4} & \textbf{84.2} & {77.8} & {87.5} & {87.7} & {81.8} & {81.0}\\
\midrule
CSS-LM-ST & {89.2} & {87.0} & \textbf{57.5} & {53.5} & {86.0} & {85.0} & {82.0} & {78.2} & {88.7} & {88.4} & \textbf{82.4} & {81.0}\\
CSS-LM & \textbf{89.5} & \textbf{87.7} & \textbf{57.5} & \textbf{54.8} & {86.0} & \textbf{85.4} & \textbf{84.2} & \textbf{78.9} & \textbf{89.0} & \textbf{89.0} & {82.3} & \textbf{81.9}\\
\midrule
\multicolumn{13}{c}{Fine-tune on few-shot setting ($K_{y}=16$)}\\
\midrule
Standard~\cite{liu2019roberta} & {68.9} & {63.9} & {39.4} & \textbf{36.1} & {74.3} & {75.0} & {45.7} & {43.7} & {46.9} & {40.5} & {47.6} & {44.1} \\
SCF~\cite{gunel2021supervised} & {69.1} & {65.3} & \textbf{39.6} & {35.5} & {75.8} & {75.0} & \textbf{50.9} & {47.1} & {52.0} & {40.3} & {46.9} & {45.4} \\
\midrule
CSS-LM-ST & {71.1} & \textbf{70.0} & {39.5} & {36.0} & {75.9} & {76.9} & {50.4} & {46.6} & {53.0} & {46.1} & {48.2} & {47.6}\\
CSS-LM & \textbf{73.0} & \textbf{70.0} & {39.5} & \textbf{36.1} & \textbf{77.5} & \textbf{77.4} & {48.8} & \textbf{47.5} & \textbf{54.7} & \textbf{47.4} & \textbf{49.0} & \textbf{48.3} \\
\bottomrule
\end{tabular}}}
\end{center}  
\end{table*}

\subsection{Datasets and Tasks}
\label{ssec:datasets}

We conduct our experiments on three typical text classification tasks including sentiment classification, intent classification, and relation extraction: 

(1)~\textbf{Sentiment classification} is the task of classifying the polarity of a given sentence. Sentiment classification is a core task of text classification. For sentiment classification, we select SemEval~\cite{DBLP:conf/acl/LiX18} and SST-5~\cite{socher-etal-2013-recursive} for our experiments.

(2)~\textbf{Intent Classification} is the task of correctly labeling a natural language utterance from a predetermined set of intents. Similar to sentiment classification, intent classification is also a core task of text classification. For intent classification, we select SciCite~\cite{Cohan2019Structural} and ACL-ARC~\cite{Cohan2019Structural} for our experiments.

(3)~\textbf{Relation Extraction} is the task of predicting attributes and relations for entities in a sentence. For example, given a sentence “Barack Obama was born in Honolulu, Hawaii.”, a relation classifier aims at predicting the relation of “bornInCity”. Relation extraction is the key component for building relational knowledge graphs, and it is of crucial significance to natural language understanding applications, such as structured search, question answering, and summarization. For relation extraction, we select SciERC~\cite{luan2018multitask} and ChemProt~\cite{Beltagy2019SciBERT} for our experiments. More details of these datasets are shown in Table~\ref{table:dataset}.

To build few-shot learning settings, we randomly sample a part of instances from the dataset as the training set $\mathcal{T}$. Additionally, we prepare the open-domain corpora $\mathcal{O}$ consisting of unused instances that share the same classes with~$\mathcal{T}$ from all downstream datasets. We also add English Wikipedia, which is used to train the original \BertBASE~\cite{devlin2018bert} and \RoBERTaBASE~\cite{liu2019roberta}, into $\mathcal{O}$. CSS-LM will leverage the open-domain corpora $\mathcal{O}$ to perform contrastive semi-supervised learning; thus, to fairly compare baseline PLMs such as \BertBASE~and \RoBERTaBASE, the baseline PLMs are all previously pre-trained on the open-domain corpora $\mathcal{O}$ in the experiments.

\subsection{Experimental Settings}
\label{ssec:experimental_settings}
We choose \BertBASE~\cite{devlin2018bert} and \RoBERTaBASE~\cite{liu2019roberta} as our encoders, using the official released parameters~\footnote{\url{https://storage.googleapis.com/bert_models/2020_02_20/uncased_L-12_H-768_A-12.zip}},\footnote{\url{https://dl.fbaipublicfiles.com/fairseq/models/roberta.base.tar.gz}}. The other parameters of CSS-LM are all initialized randomly. 

For training, we set the learning rate as $2 \times 10^{-5}$, the batch size as $4$. The remaining settings follow the original ones of \BertBASE~and \RoBERTaBASE. For the retrieved instance size, we perform a grid search over multiple hyper-parameters $\{16,32,48,64\}$, and take the best one measured on the whole development set for our model.

The objective function of CSS-LM is $\mathcal{L} = \mathcal{L}_{CE} + \mathcal{L}_{CSS} + \mathcal{L}_D + \mathcal{L}_C$, which enables CSS-LM to retrieve task-related instances and enhance the performance of downstream tasks; however, $\mathcal{L}_D$ and $\mathcal{L}_C$ terms may make the performance drop on some downstream tasks during contrastive semi-supervised learning sometimes. Thus, we degrade the objective function $\mathcal{L} = \mathcal{L}_{CE} + \mathcal{L}_{CSS} + \mathcal{L}_D + \mathcal{L}_C$ to $\mathcal{L} = \mathcal{L}_{CE} + \mathcal{L}_{CSS}$ when the performance starts to drop in the development sets. Then, we continuously train CSS-LM with the new objective function.

Besides, to avoid the result instability~\cite{dodge2020finetuning,zhang2020revisiting}, we report the average performance across $5$ different randomly sampled data splits.

\begin{figure*}[!t]
\centering
\subfigure[SemEval~(\RoBERTaBASE)]{
\begin{minipage}[t]{0.32\linewidth}
\centering
\includegraphics[width=0.95\textwidth]{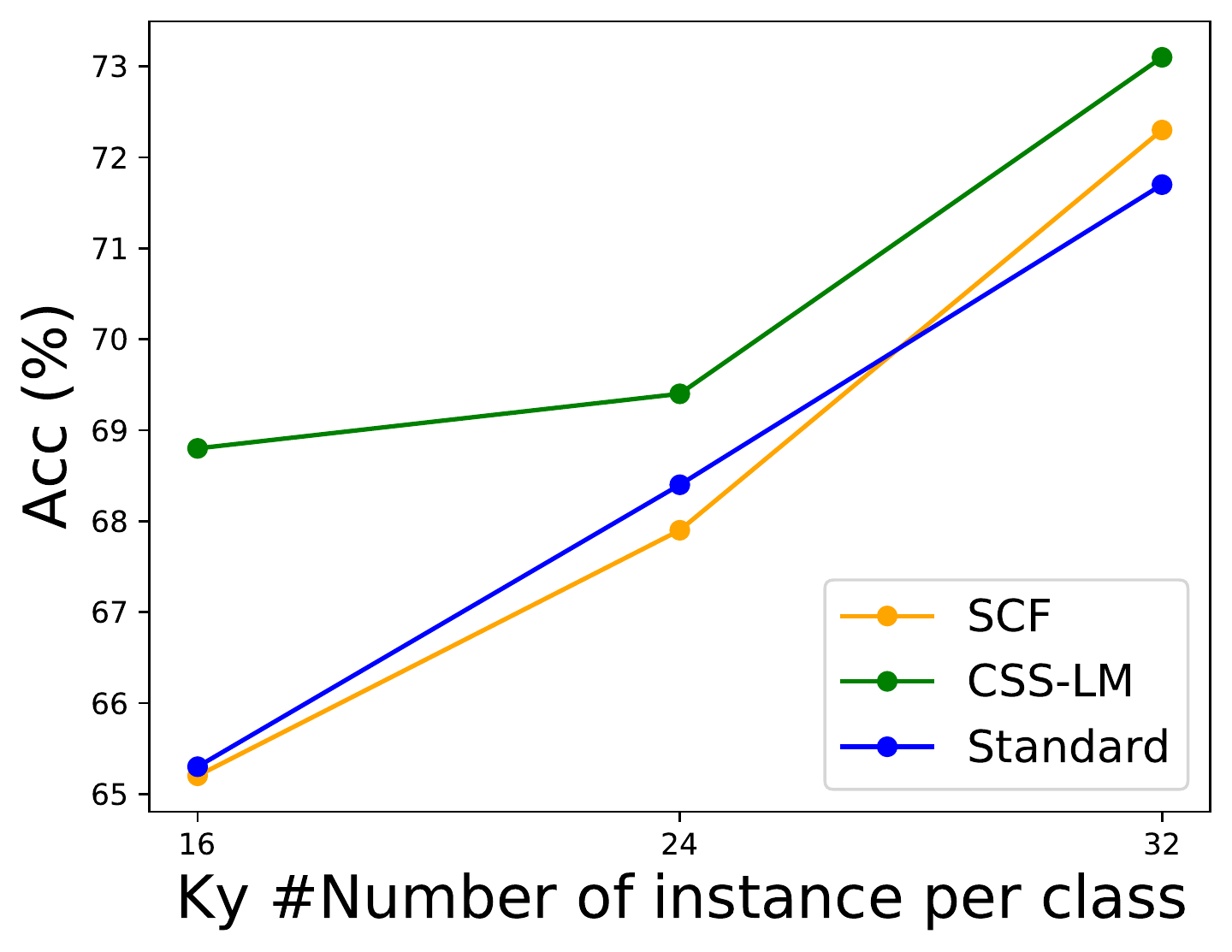}
\end{minipage}%
}%
\subfigure[SciCite~(\RoBERTaBASE)]{
\begin{minipage}[t]{0.32\linewidth}
\centering
\includegraphics[width=0.95\textwidth]{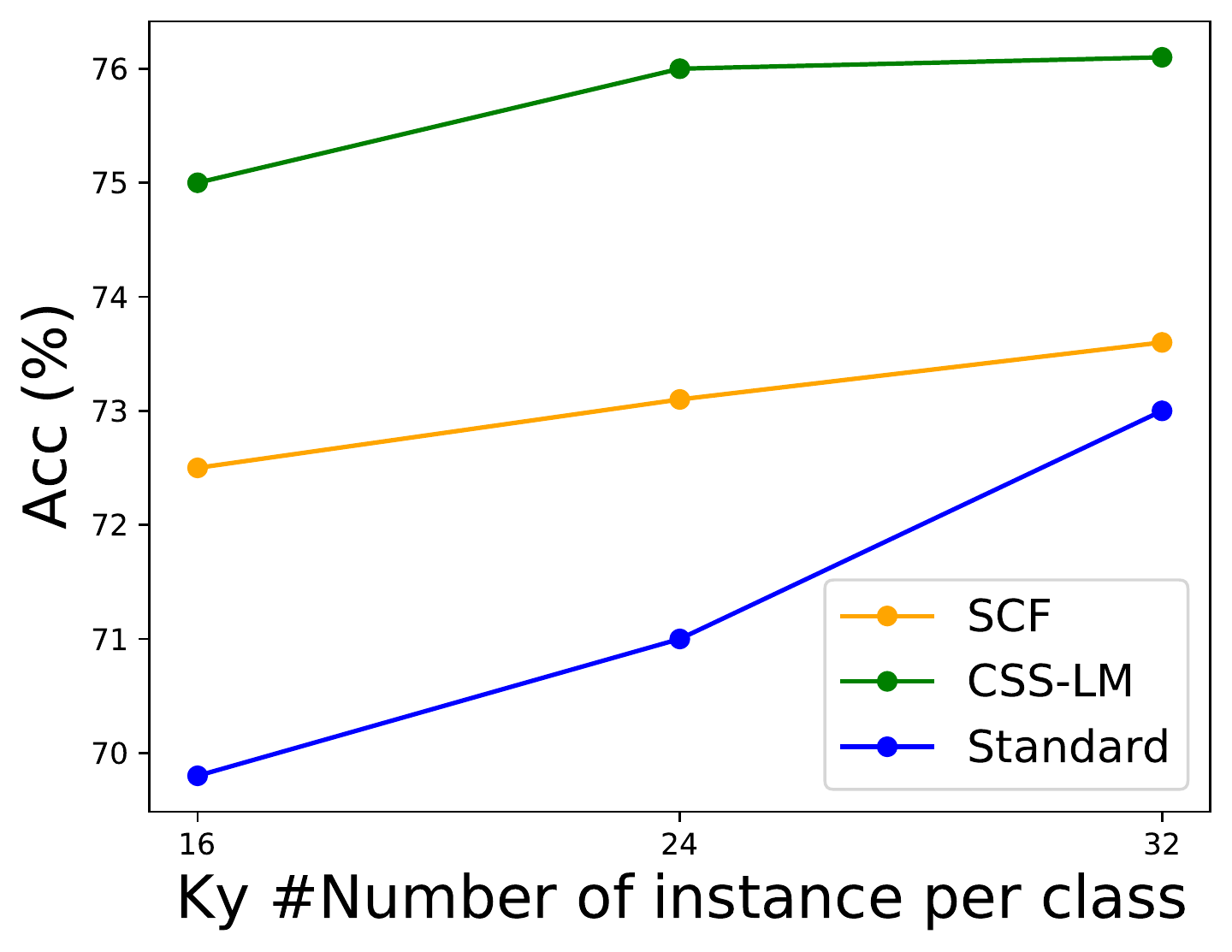}
\end{minipage}%
}%
\subfigure[SciERC~(\RoBERTaBASE)]{
\begin{minipage}[t]{0.32\linewidth}
\centering
\includegraphics[width=0.95\textwidth]{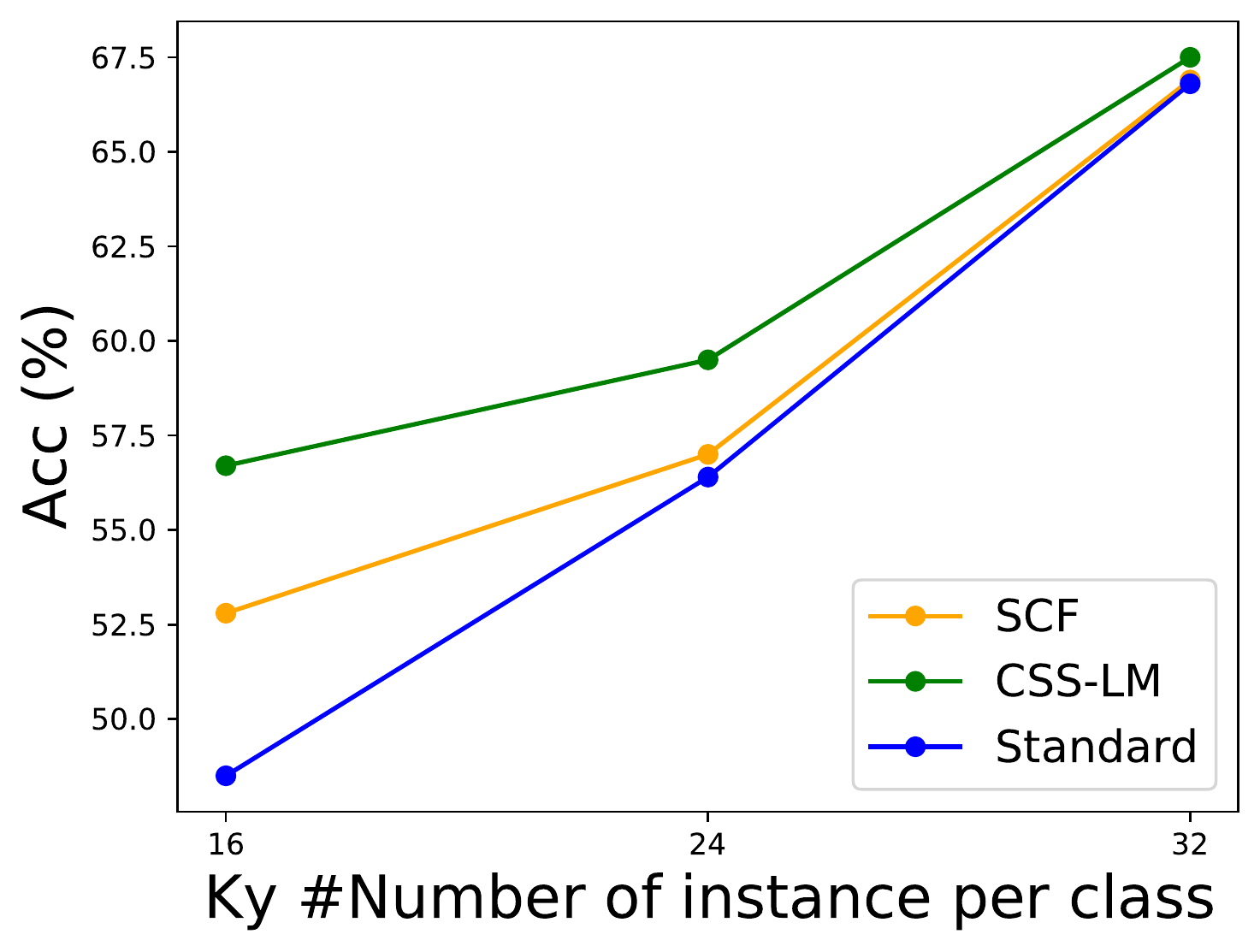}
\end{minipage}%
}%

\subfigure[SemEval~(\BertBASE)]{
\begin{minipage}[t]{0.32\linewidth}
\centering
\includegraphics[width=0.95\textwidth]{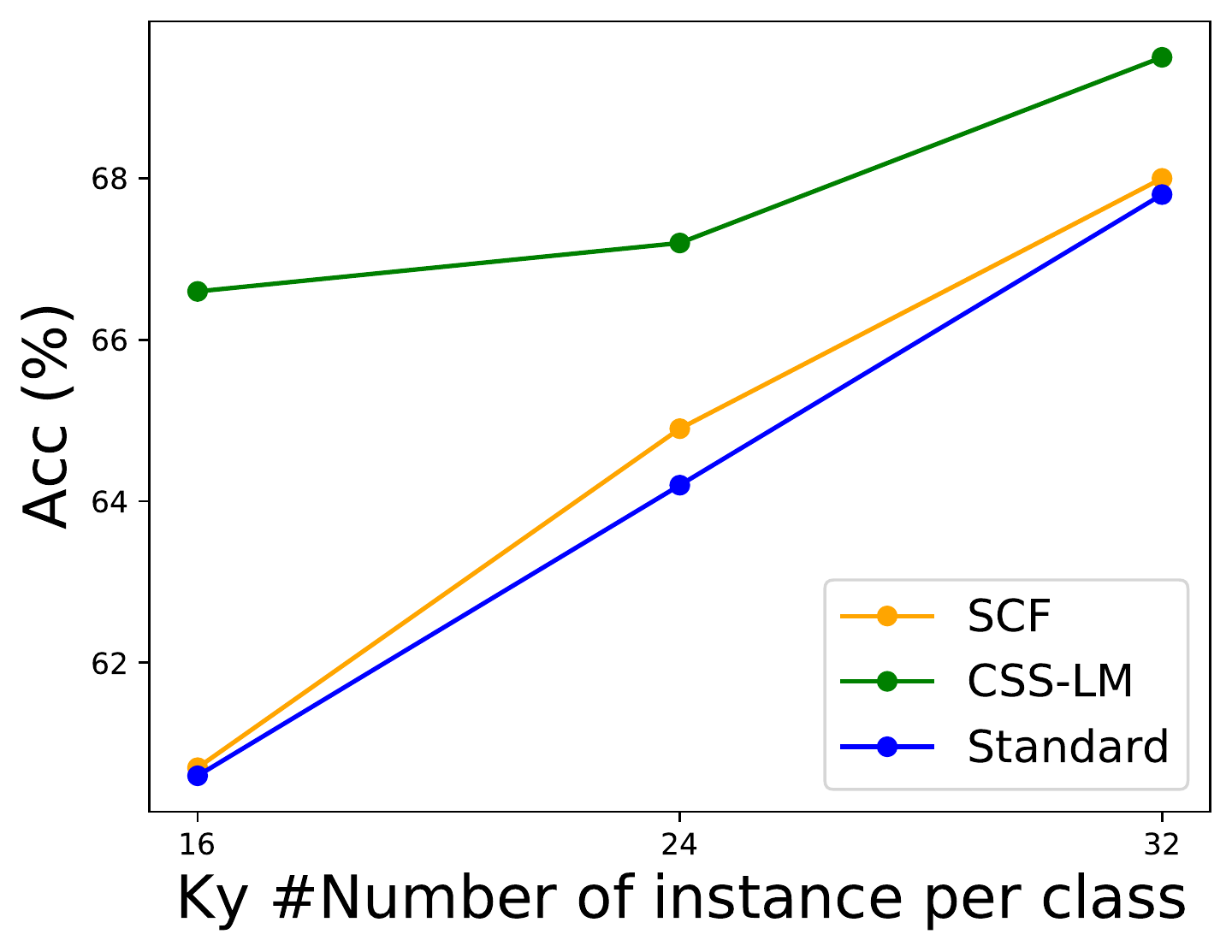}
\end{minipage}%
}%
\subfigure[SciCite~(\BertBASE)]{
\begin{minipage}[t]{0.32\linewidth}
\centering
\includegraphics[width=0.95\textwidth]{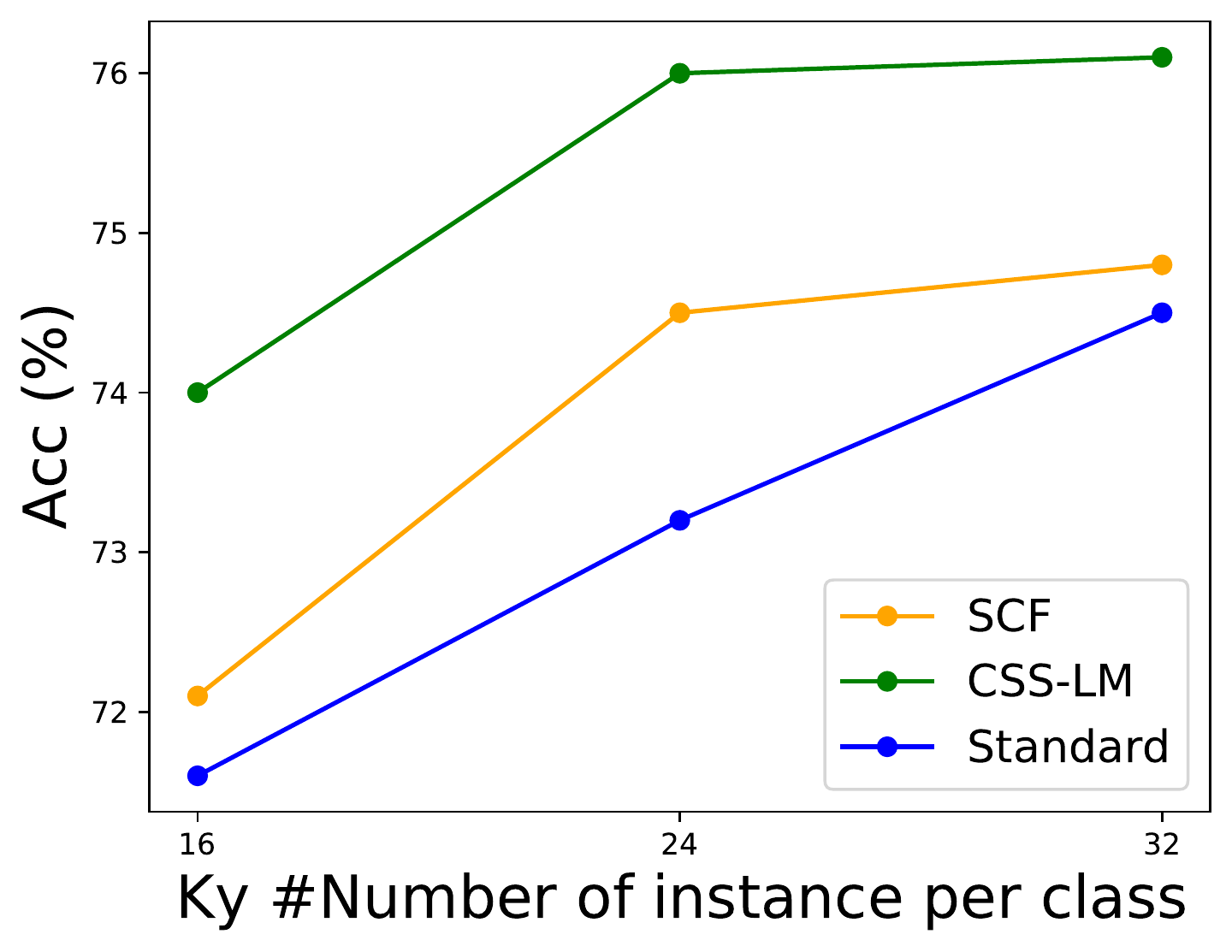}
\end{minipage}%
}%
\subfigure[SciERC~(\BertBASE)]{
\begin{minipage}[t]{0.32\linewidth}
\centering
\includegraphics[width=0.95\textwidth]{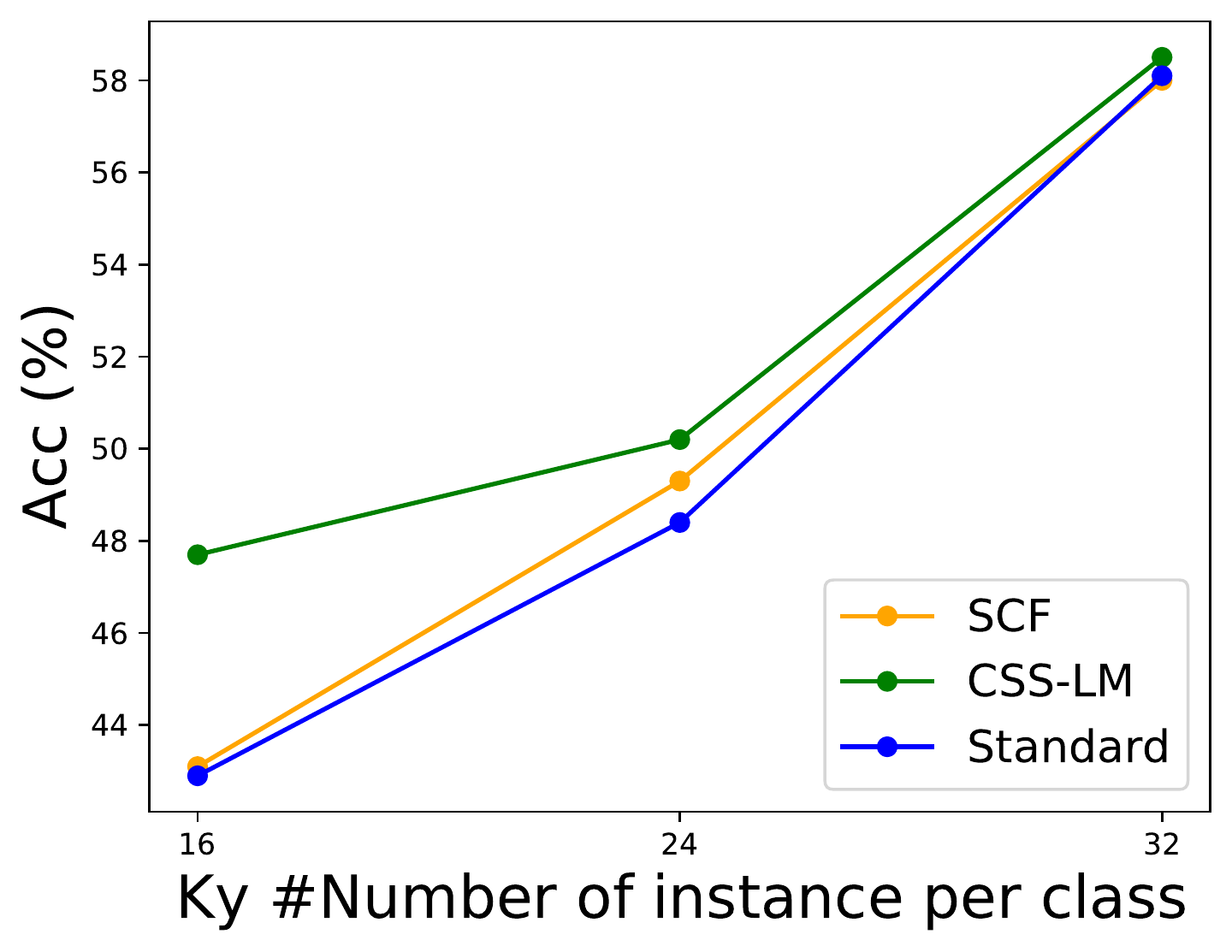}
\end{minipage}%
}%
\caption{The results (\%) of different fine-tuning strategies on the development sets of SemEval, Scicite and SciERC, with different numbers of instance per class.}
\label{fig:number_of_k}
\end{figure*}

\begin{figure}[t]
\centering
    \subfigure[\RoBERTaBASE]{
    \begin{minipage}[t]{0.9\linewidth}
    \centering
    \includegraphics[width=1\textwidth]{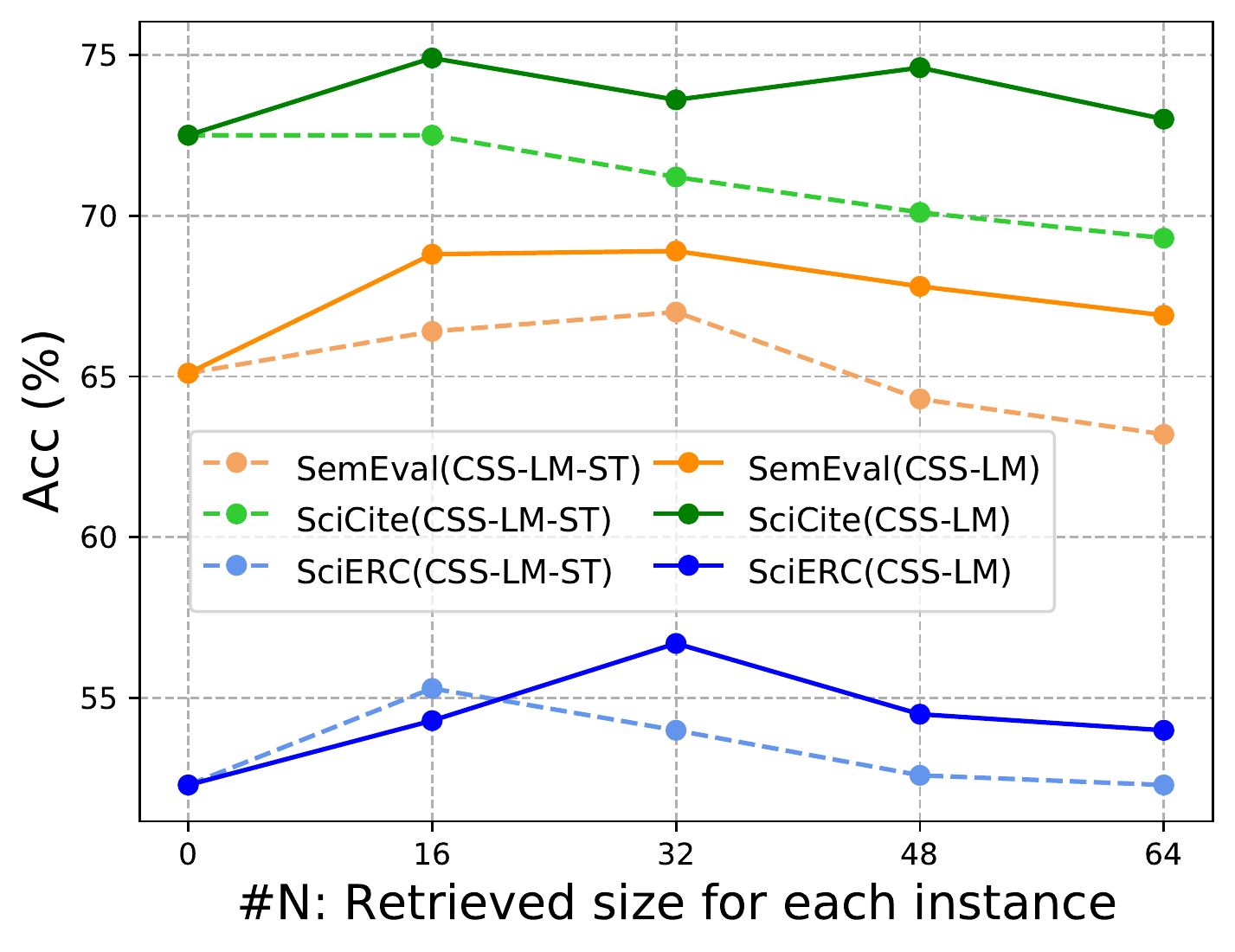}
    \end{minipage}%
    }%
    
    \subfigure[\BertBASE]{
    \begin{minipage}[t]{0.9\linewidth}
    \centering
    \includegraphics[width=1\textwidth]{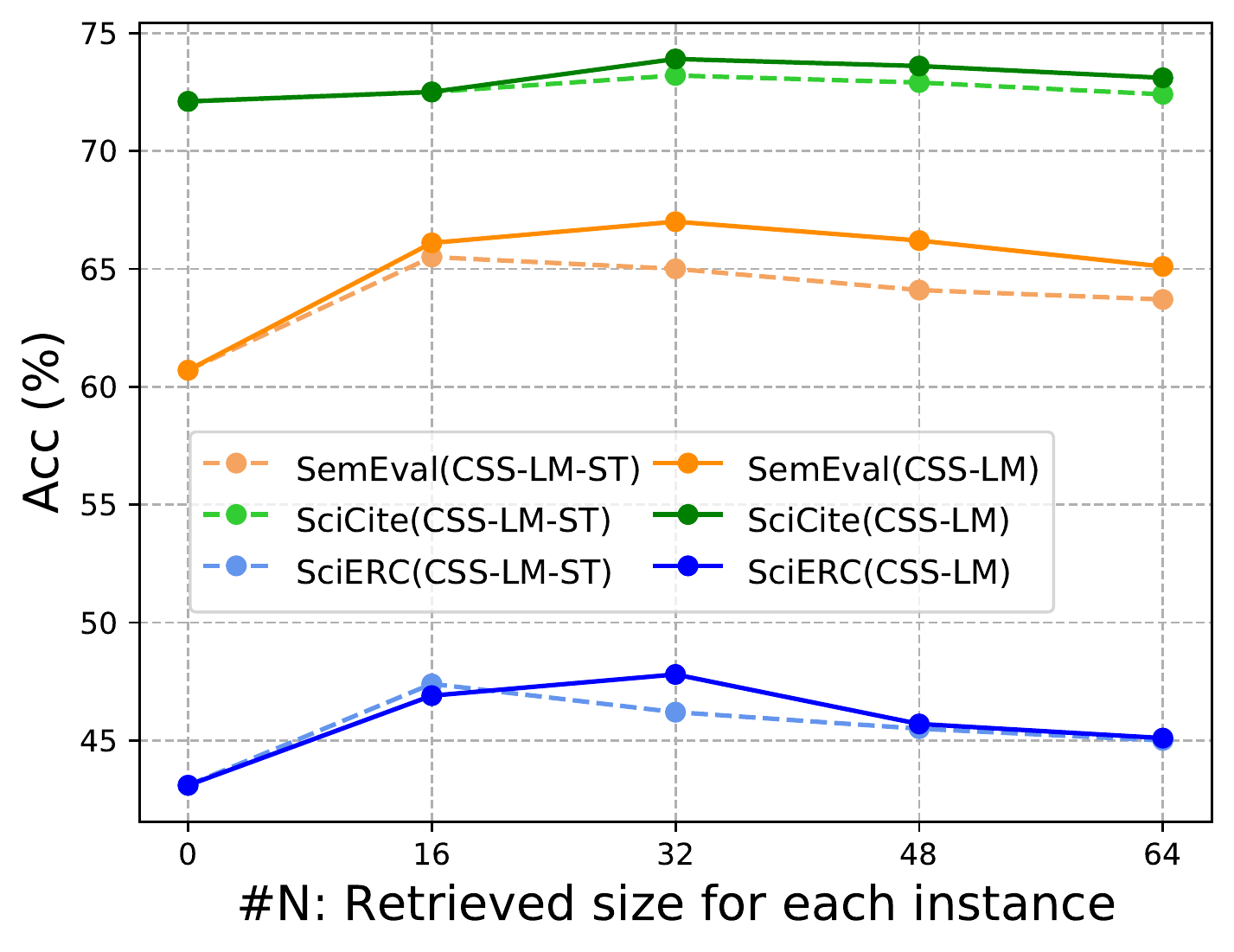}
    \end{minipage}%
    }%
\caption{The effect (\%) of retrieved instance size on CSS-LM and CSS-LM-ST on the development sets of SemEval, Scicite and SciERC.}
\label{fig:size_of_retrieve}
\end{figure}

\subsection{Baselines}
\label{ssec:baseline}

We compare our CSS-LM with the following effective fine-tuning strategies: 

\textbf{Standard fine-tuning~(Standard)}, which is the typical fine-tuning method in PLMs~\cite{devlin2018bert,liu2019roberta} and its training objective is $\mathcal{L}_{CE}$ mentioned in Eq.~(\ref{eq:loss}).

\textbf{Supervised contrastive fine-tuning~(SCF)} \cite{gunel2021supervised}, which only performs supervised contrastive learning with the supervised data of downsteam tasks, and its fine-tuning loss is $\mathcal{L}_{CE} + \mathcal{L}_{CS}$, referring to Eq.~(\ref{equ:sc}) and Eq.~(\ref{eq:loss}).

\textbf{CSS-LM-ST}, a variant of our CSS, which also retrieves task-related in-domain data with domain-related and class-related relatedness (i.e., the positive instances) by contrastive semi-supervised learning as we mentioned in the section~\ref{ssec:contrastive_semisupervised_learning}. The difference is CSS-LM-ST performing the standard fine-tuning method with pseudo labeling~\cite{lee2013pseudo}, which is the simple and efficient semi-supervised learning method for deep neural networks, instead of contrastive semi-supervised learning to capture critical features from the retrieved instances. We can denote the learning objective of CSS-LM-ST as $\mathcal{L}_{D} + \mathcal{L}_{C} + \mathcal{L}_{CE}^{'}$, where $\mathcal{L}_{CE}^{'}$ is the same downstream task fine-tuning loss in the section~\ref{ssec:downstream_task_finetuning_and_optimization} but leverages extra pseudo labeled instances of open-domain corpora. 

\textbf{CSS-LM} and \textbf{CSS-LM-ST} leverage the same method to retrieve task-related instances but apply different mechanisms to learn semantic features from the retrieved instances. CSS-LM learns semantic features by contrastive semi-supervised learning. CSS-LM-ST uses the conventional semi-supervised learning with pseudo labels to learn the features, which may be sensitive to the quality of labels. The key advantage of CSS-LM is free to pseudo labels. In the following parts, we will extensively study CSS-LM and CSS-LM-ST.

\subsection{Overall Results}
\label{ssec:NLU_task}

We conduct experiments on the three selected tasks under the standard and few-shot settings ($K_{y}=16$). The results are shown in Table~\ref{table:NLU_task_table}. From the table, we can see that:

(1) Our CSS-LM framework achieves improvements on almost all six datasets compared to the baseline models (including the state-of-the-art SCF), especially under the few-shot settings. This demonstrates that our contrastive semi-supervised framework for fine-tuning could effectively capture the important semantic features for the task from the large-scale unlabeled data.

(2) Although utilizing the retrieved task-related in-domain data can help fine-tuning, 
CSS-LM-ST outperforms standard fine-tuning yet does not obtain the same performance improvements as CSS-LM under the few-shot settings. It indicates that our framework can retrieve high-quality instances by contrastive semi-supervised learning; applying contrastive semi-supervised learning is better than assigning pseudo labels to the retrieved instances to train classifiers. In fact, under the few-shot settings, some retrieved instances will be linked to some classes close to their implicit golden labels rather than the golden labels. Therefore, directly annotating pseudo labels may lead to a biased model. In contrast, these instances corresponding to wrong classes can still provide correlation information for the contrastive learning of CSS-LM.

(3) We also compare CSS-LM under two different settings: fine-tuning with the few-shot instances and the whole training set instances. Although CSS-LM outperforms most baseline models under the few-shot setting, CSS-LM obtains slight improvements when fine-tuning with sufficient training data. This is an intuitive observation that directly fine-tuning PLMs will work well when the amount of data is sufficient. However, not all NLP tasks have enough data; low-resource tasks and long-tail classes are very common. Our framework can improve the model performance in few-shot learning scenarios without weakening the model performance.

\subsection{Effect of Supervised Data Size}

In this part, we explore the effect of the supervised instance number $K_{y}$ for each class. From Fig.~\ref{fig:number_of_k}, we have the following findings: 

(1) Under the few-shot settings, both SCF and CSS-LM outperform the standard fine-tuning strategy on two datasets consistently. It indicates that performing contrastive learning between instances of different classes could help extract informative semantic features to distinguish them, and benefit downstream tasks in the fine-tuning stage. 

(2) CSS-LM has better results than SCF on all the datasets of our experiments, especially when the supervised data size is small. It demonstrates that although SCF could discover the class relatedness from the supervised data to some extent, it may ignore semantic information that is trivial in the limited supervised data but crucial for the unseen task. In contrast, our CSS-LM could effectively take account of the unlabeled data to capture the informative semantic features not expressed by supervised data.

\subsection{Effect of Retrieved Instance Size}

In order to learn discriminative features, CSS-LM needs to leverage proper positive and negative pairs. As for CSS-LM-ST, pseudo labeling quality is essential for the performance. In this part, we explore the effect of different retrieved instance size on CSS-LM and CSS-LM-ST. As shown in Fig.~\ref{fig:size_of_retrieve}, we have the following findings:

(1)~When the size is small, CSS-LM retrieves too similar instances to make them close to each other, according to the contrastive learning paradigm, which cannot learn the essential information. As the size increases, the model performance can gradually increase. Since more informative instances are retrieved from open-domain corpora, CSS-LM will take unrelated instances as similar instances to learn the false information when the size is too large. In future, it is meaningful to study how to denoise the retrieved instances for semi-supervised fine-tuning.

(2)~When the size becomes large, the performance of CSS-LM-ST decays earlier than CSS-LM since CSS-LM-ST needs to learn features from high-quality labeled instances. Instead of learning the features from an individual instance, CSS-LM learns by comparing among different instances. Therefore, CSS-LM can better leverage sufficient unlabeled instances.

\begin{table}[t]
\caption{\label{table:in_out_domain} The results (\%) of CSS-LM on the open-domain corpora and in-domain corpora. As in-domain corpora do not consider domain-level semantics, we perform CSS-LM$_\texttt{[C]}$ on in-domain corpora.}
\begin{center} 
\small
{
\setlength{\tabcolsep}{1pt}{
\begin{tabular}{l|l|l|l}
\toprule
  \multicolumn{4}{c}{Fine-tune on few-shot $K_{y}=16$}\\
  \midrule
  \textbf{Task} & \multicolumn{1}{c|}{\textbf{\makecell[c]{Sentiment \\ Classification}}} & \multicolumn{1}{c|}{\textbf{\makecell[c]{Intent \\ Classification}}} & \multicolumn{1}{c}{\textbf{\makecell[c]{Relation \\ Classification}}}\\
  \midrule
  \textbf{Dataset} & \multicolumn{1}{c|}{\textbf{SemEval}} & \multicolumn{1}{c|}{\textbf{SciCite}} & \multicolumn{1}{c}{\textbf{SciERC}}
  \\
\midrule \midrule
\multicolumn{4}{c}{\RoBERTaBASE}\\
\midrule
\multicolumn{4}{c}{Open-domain corpora}\\
\midrule
CSS-LM$_\texttt{[D+C]}$ & \multicolumn{1}{c|}{\textbf{73.0}} & \multicolumn{1}{c|}{\textbf{77.5}} & \multicolumn{1}{c}{54.7} \\
\midrule
\multicolumn{4}{c}{In-domain corpora}\\
\midrule
CSS-LM$_\texttt{[C]}$ & \multicolumn{1}{c|}{\textbf{73.0}} & \multicolumn{1}{c|}{75.9} & \multicolumn{1}{c}{\textbf{54.9}} \\
\midrule\midrule
\multicolumn{4}{c}{\BertBASE}\\
\midrule
\multicolumn{4}{c}{Open-domain corpora}\\
\midrule
CSS-LM$_\texttt{[D+C]}$ & \multicolumn{1}{c|}{70.0} & \multicolumn{1}{c|}{\textbf{77.4}} & \multicolumn{1}{c}{47.1} \\
\midrule
\multicolumn{4}{c}{In-domain corpora}\\
\midrule
CSS-LM$_\texttt{[C]}$ & \multicolumn{1}{c|}{\textbf{71.0}} & \multicolumn{1}{c|}{76.1} & \multicolumn{1}{c}{\textbf{47.4}} \\
\bottomrule
\end{tabular}}}
\end{center}  
\end{table}

\begin{table}[h]
\caption{\label{table:domain_class_term} The results (\%) of CSS-LM utilizing semantic relatedness at different levels.}
\begin{center} 
\small
{
\setlength{\tabcolsep}{1pt}{
\begin{tabular}{l|l|l|l}
\toprule
  \multicolumn{4}{c}{Fine-tune on few-shot $K_{y}=16$ in open-domain corpora}\\
  \midrule
  \textbf{Task} & \multicolumn{1}{c|}{\textbf{\makecell[c]{Sentiment \\ Classification}}} & \multicolumn{1}{c|}{\textbf{\makecell[c]{Intent \\ Classification}}} & \multicolumn{1}{c}{\textbf{\makecell[c]{Relation \\ Classification}}}\\
  \midrule
  \textbf{Dataset} & \multicolumn{1}{c|}{\textbf{SemEval}} & \multicolumn{1}{c|}{\textbf{SciCite}} & \multicolumn{1}{c}{\textbf{SciERC}}
  \\
\midrule
\midrule
\multicolumn{4}{c}{\RoBERTaBASE}\\
\midrule
CSS-LM$_\texttt{[C]}$ & \multicolumn{1}{c|}{\textbf{73.6}} & \multicolumn{1}{c|}{75.1} & \multicolumn{1}{c}{53.6} \\
CSS-LM$_\texttt{[D]}$ & \multicolumn{1}{c|}{72.1} & \multicolumn{1}{c|}{75.7} & \multicolumn{1}{c}{51.9} \\
CSS-LM$_\texttt{[D+C]}$ & \multicolumn{1}{c|}{73.0} & \multicolumn{1}{c|}{\textbf{77.5}} & \multicolumn{1}{c}{\textbf{54.7}} \\

\midrule\midrule
\multicolumn{4}{c}{\BertBASE}\\
\midrule
CSS-LM$_\texttt{[C]}$ & \multicolumn{1}{c|}{69.8} & \multicolumn{1}{c|}{75.8} & \multicolumn{1}{c}{46.0} \\
CSS-LM$_\texttt{[D]}$ & \multicolumn{1}{c|}{68.1} & \multicolumn{1}{c|}{74.4} & \multicolumn{1}{c}{45.7} \\
CSS-LM$_\texttt{[D+C]}$ & \multicolumn{1}{c|}{\textbf{70.0}} & \multicolumn{1}{c|}{\textbf{77.4}} & \multicolumn{1}{c}{\textbf{47.7}} \\
\bottomrule
\end{tabular}}}
\end{center}  
\end{table}

\subsection{Difference between Retrieving Instances from In-domain Data and Open-domain Data}

To show our framework can automatically extract in-domain instances from open-domain corpora, we perform CSS-LM$_\texttt{[D+C]}$ for the open-domain corpora, and perform CSS-LM$_\texttt{[C]}$ for the in-domain corpora. Specifically, the in-domain corpora contain some unused instances share same classes with $\mathcal{T}$. CSS-LM$_\texttt{[D]}$ is the domain-level term of our CSS-LM and CSS-LM$_\texttt{[C]}$ is the class-level term, CSS-LM$_\texttt{[D+C]}$ is the combination of $_\texttt{[D]}$ and $_\texttt{[C]}$. The results are shown in Table~\ref{table:in_out_domain}. 

From the table, we can see that:
CSS-LM$_\texttt{[D+C]}$ can achieve comparable performance with CSS-LM$_\texttt{[C]}$ trained on the in-domain corpora. Although directly retrieving in-domain data is easier than retrieving open-domain data, we can see CSS-LM$_\texttt{[C]}$ is only slightly better than CSS-LM$_\texttt{[D+C]}$ in the well-defined domain such as SemEval~(Restaurant review) and ChemProt~(Biology); CSS-LM$_\texttt{[D+C]}$ even outperforms CSS-LM$_\texttt{[C]}$ in SciCite, which belongs to multiple domains (not well-defined). It demonstrates that CSS-LM can efficiently learn to distinguish coarse-grained domains from a large amount of the open-domain corpora, free to domain dependence.

\subsection{Contribution of Domain-Related and Class-Related Semantics to CSS-LM}
As we empirically apply domain-related and class-related semantics for our framework, we wonder which semantic levels makes the greater contribution to our framework. In this part, we study the effect of the domain-level and class-level terms of CSS-LM as shown in Table~\ref{table:domain_class_term}. 

From the results, we can find that: CSS-LM$_\texttt{[D+C]}$ is comparable to CSS-LM$_\texttt{[C]}$ and outperforms CSS-LM$_\texttt{[D]}$ in almost all datasets. Therefore, compared with the domain-level term, the class-level term is essential to retrieving high-quality instances to enhance CSS-LM performance.  

Although this paper mainly focuses on applying the contrastive semi-supervised framework for fine-tuning rather than exploring retrieving informative instances, we think how to better consider domain-level semantic remains an interesting problem in the future.

\begin{figure*}[t]
\centering
\subfigure[Stardard (Whole Set)]{
\begin{minipage}[t]{0.24\linewidth}
\centering
\includegraphics[width=0.9\textwidth]{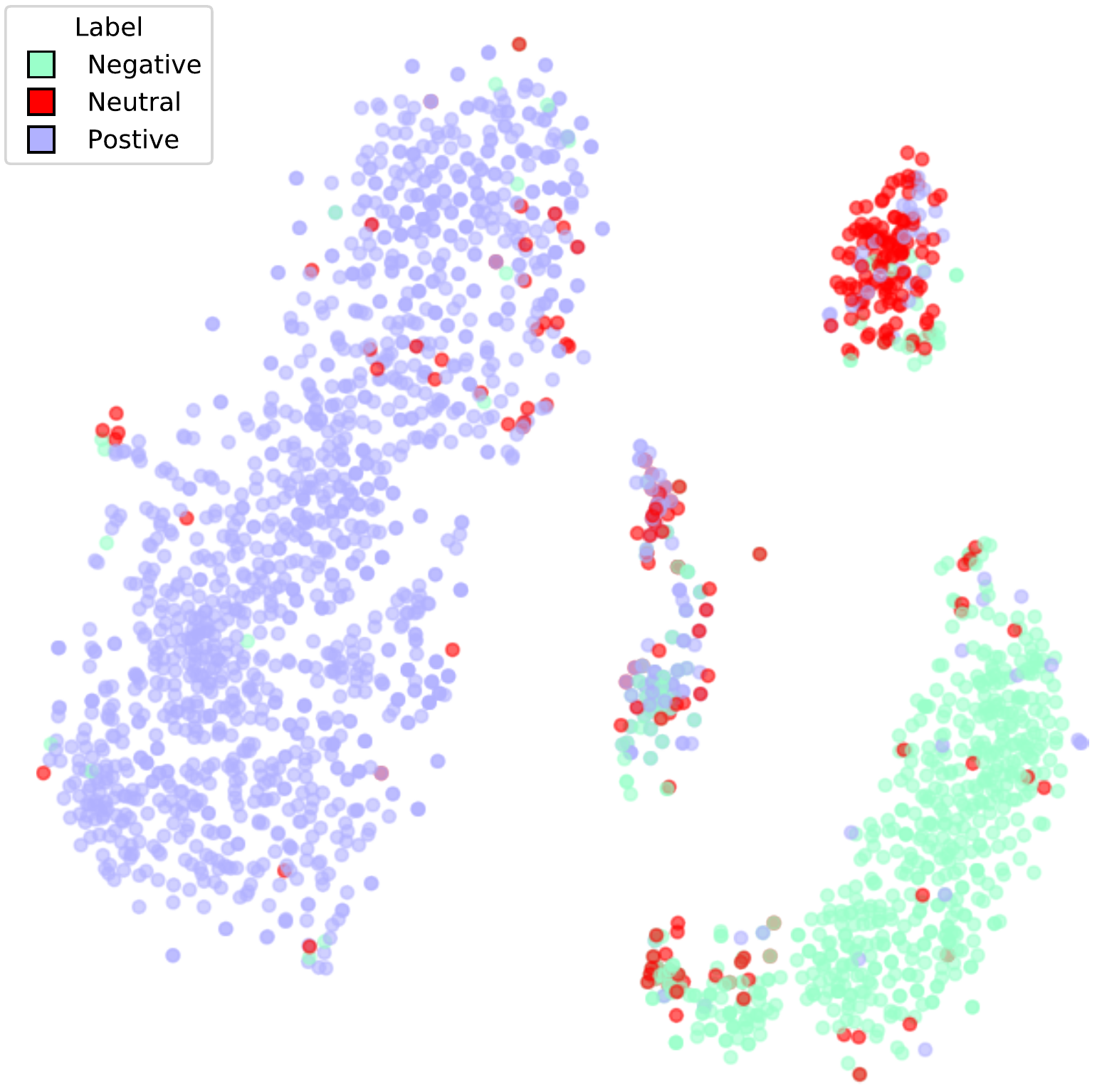}
\end{minipage}%
}%
\subfigure[Stardard]{
\begin{minipage}[t]{0.24\linewidth}
\centering
\includegraphics[width=0.9\textwidth]{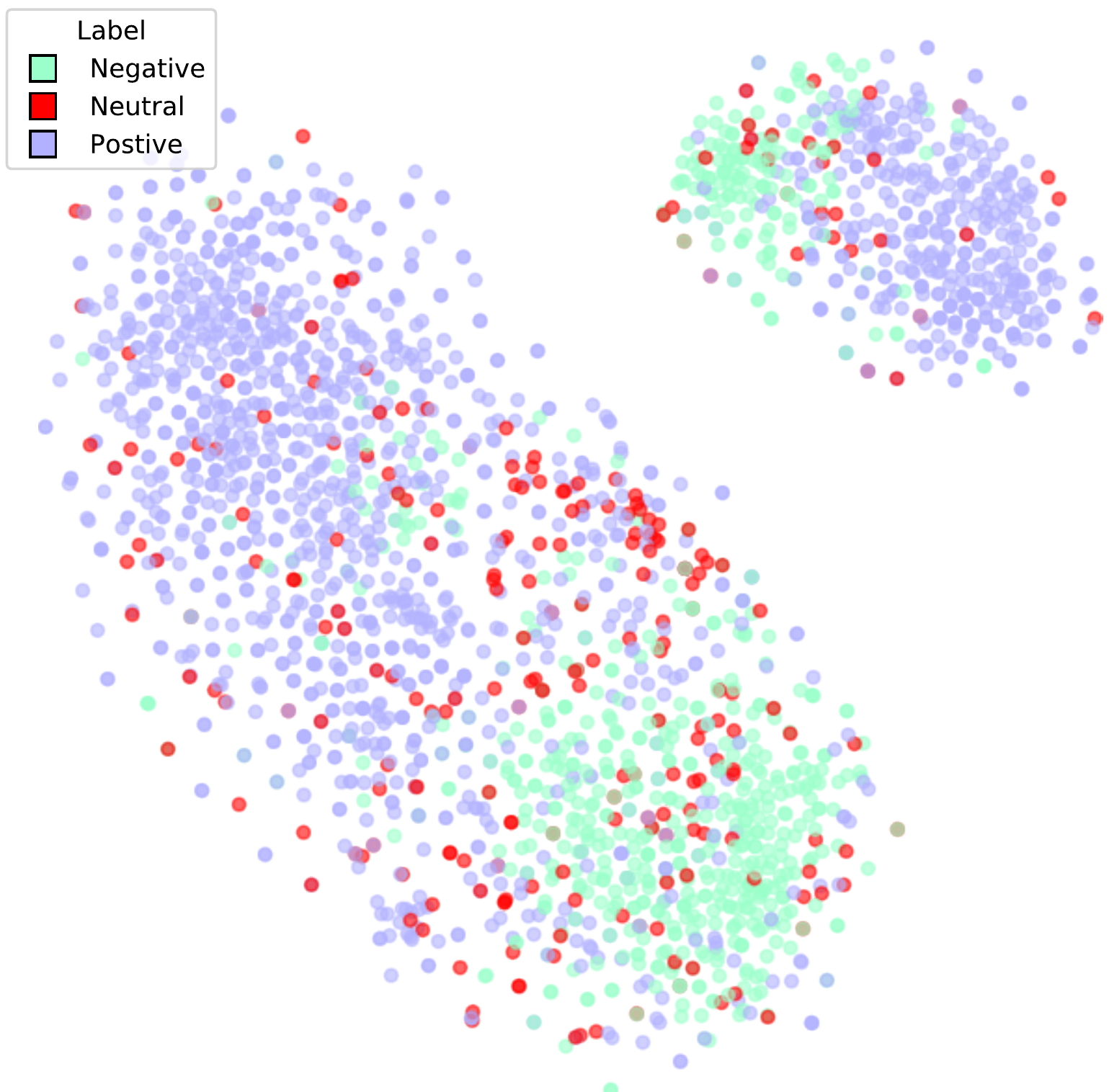}
\end{minipage}%
}%
\subfigure[SCF]{
\begin{minipage}[t]{0.24\linewidth}
\centering
\includegraphics[width=0.9\textwidth]{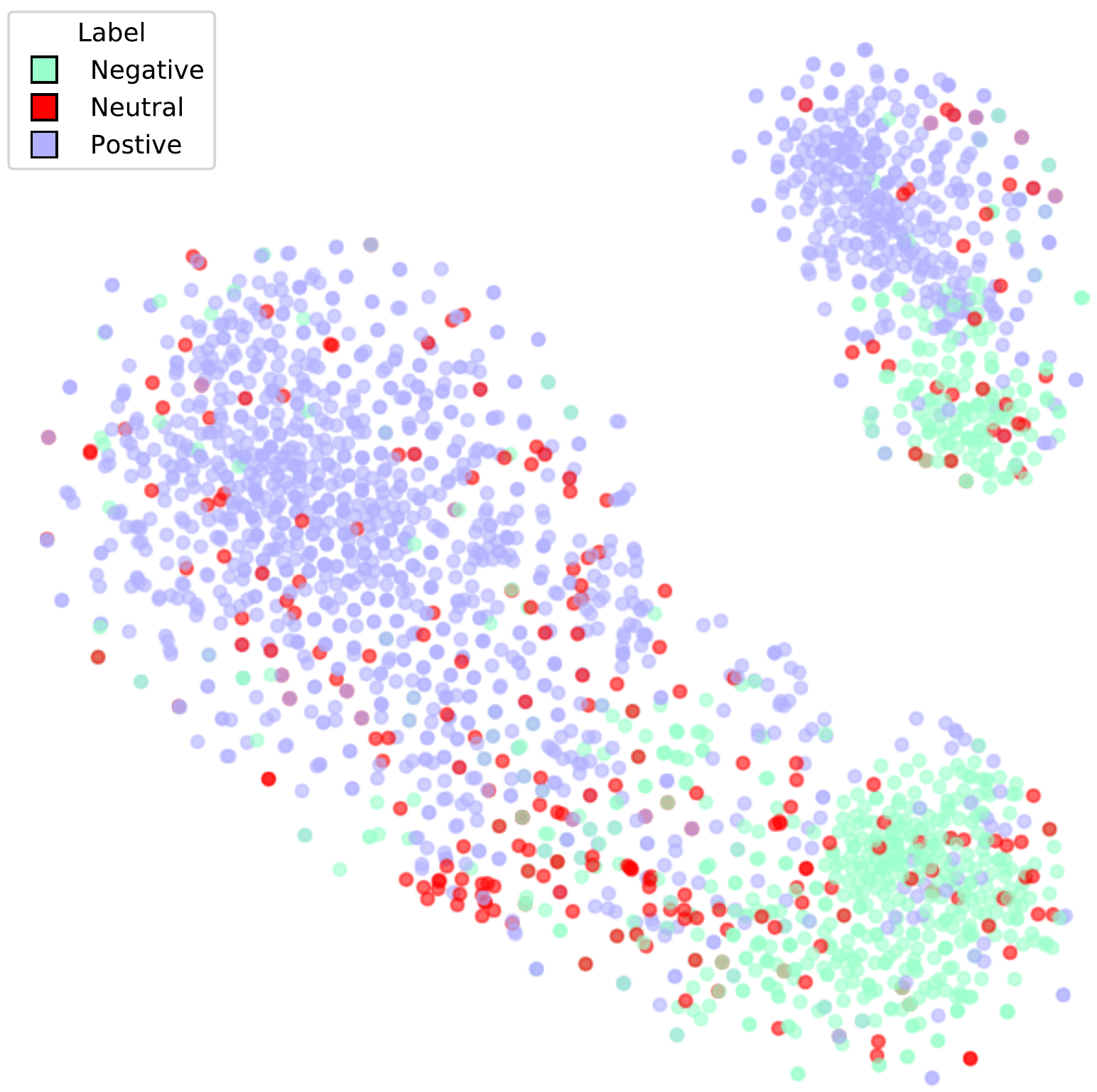}
\end{minipage}
}%
\subfigure[CSS-LM]{
\begin{minipage}[t]{0.24\linewidth}
\centering
\includegraphics[width=0.9\textwidth]{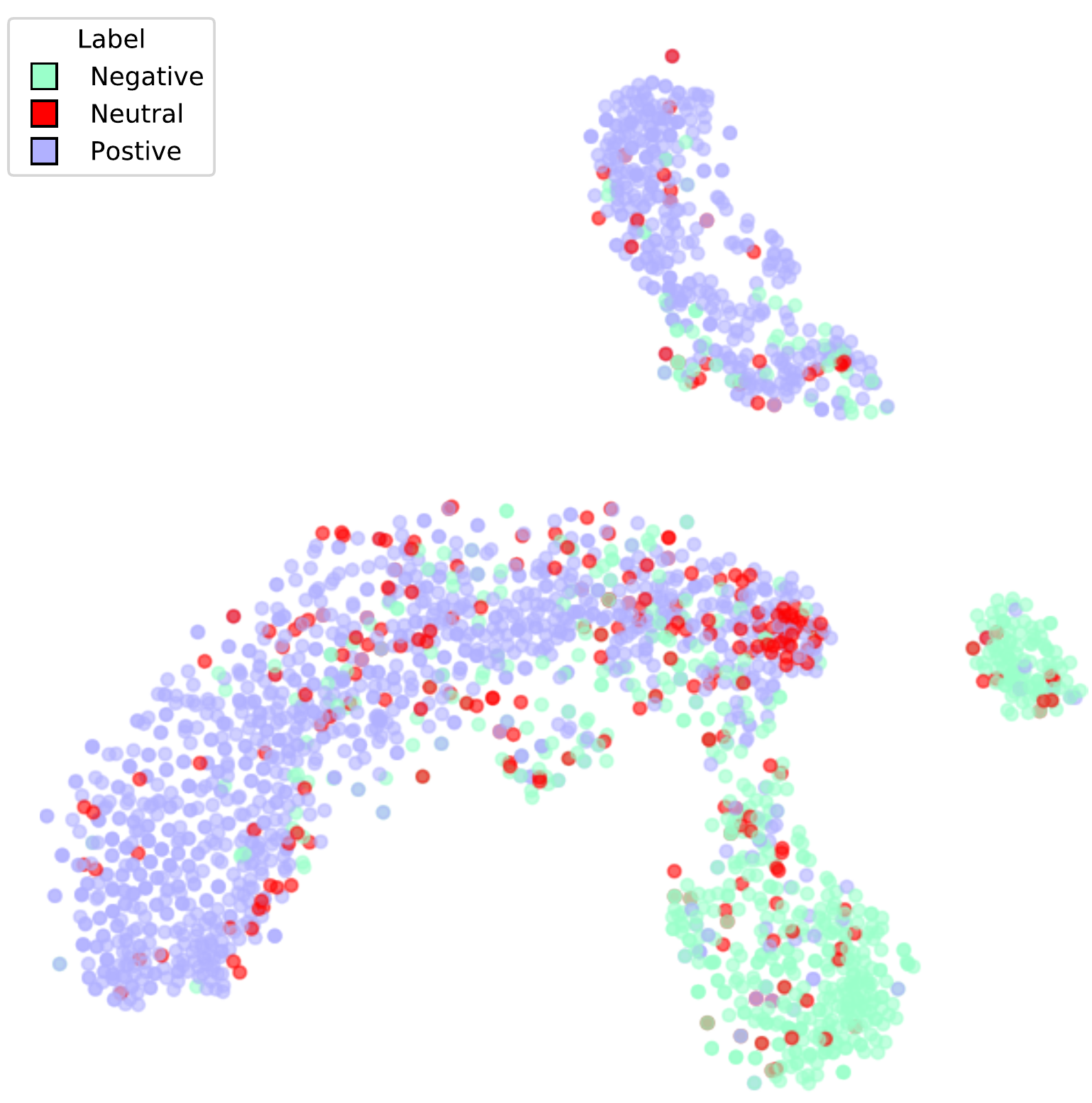}
\end{minipage}
}%
\caption{The tSNE plots of the embeddings learned by standard, SCF, and CSS-LM fine-tuning methods on the SemEval dataset. Green: Negative emotion; Blue: Positive emotion; Red: Neutral emotion. Except the embeddings in (a) are trained by the whole training set, the embeddings in (b) (c) (d) are only trained by $K_y=16$.}
\label{fig:fine-tuning_embedding}
\end{figure*}

\begin{table*}[t]
\caption{\label{table:retrieve_instances} The analyses of some retrieved instances from the corpora by CSS-LM. Yellow: Class-related; Gray: Domain-related.}
\begin{center} 
\scalebox{0.95}
{
\setlength{\tabcolsep}{1pt}{
\begin{tabular}{c|c|c}
\toprule
{Class} & {Instances} &{Related Instances} \\
\midrule\midrule
{\textbf{Negative}} & {\small{\makecell[l]{\colorbox{lightgray}{Service} was just ok, it is\colorbox{yellow}{not what you'd expect}for \$500.}}} & {\small{\makecell[l]{\colorbox{yellow}{I'm not even going to bother to describe it; speaks for itself}.\\ It's only \$1.95 for a regular slice and 4.00 for a slice with a \\\colorbox{lightgray}{mushroom}, not \colorbox{lightgray}{mushrooms}. \\}}}\\
\midrule
{\textbf{Neutral}} & {\small{\makecell[l]{A great way to make some money is to buy a case of\\ \colorbox{lightgray}{snapple} from costco and sell it right outside for only \$2.50.}}} & {\small{\makecell[l]{Try the times square cocktail -- ginger lemonade with vodka.\\
I had the \colorbox{lightgray}{tuna tartare} with sake, \colorbox{lightgray}{mushroom} ravioli with\\ pinot noir, and the \colorbox{lightgray}{chocolate} sampler $[...]$.\\ }}}\\
\midrule
{\textbf{Positive}} & {\small{\makecell[l]{The \colorbox{lightgray}{ambience} was so fun, and the prices were\colorbox{yellow}{great}, \\on top of the fact that the \colorbox{lightgray}{food} was \colorbox{yellow}{really tasty}.}}} & {\small{\makecell[l]{Overall, I would \colorbox{yellow}{highly recommend} giving this one a try. \\
$[...]$ Liverpool boss Klopp says win \colorbox{yellow}{feels perfect}. \\ 
Oh yeah ever on the west side try there sister \colorbox{lightgray}{restaurant}\\ \colorbox{lightgray}{arties cafe}.}}} \\
\bottomrule
\end{tabular}}}
\end{center}  
\end{table*}

\subsection{Visualization}
\label{sec:visulization}
We apply t-SNE~\cite{Ulyanov2016} to visualize instance embeddings of the SemEval test set learned by standard, SCF, and CSS-LM fine-tuning strategies in Fig.~\ref{fig:fine-tuning_embedding}. We give the two-dimensional points with different colors to represent its corresponding label in the downstream task. 

From the figure, we can intuitively see that direct fine-tuning with sufficient data Fig.~\ref{fig:fine-tuning_embedding}(a) has the best boundaries among all classes. Without sufficient data, all fine-tuning methods cannot detect neutral instances, but our CSS-LM can obtain the coarse-grained neutral cluster and achieve a better decision boundary between positive and negative instances.

\subsection{Case Study}
\label{sec:case_study}
As shown in Table~\ref{table:retrieve_instances}, we also give a case study to investigate whether CSS-LM can capture domain-level and class-level relatedness of instances. We color the sub-sequence containing human-annotated class-level information with yellow and domain-level information with gray.

From the table, we can see that CSS-LM can retrieve instances highly agreed with humans in most cases. Interestingly, the neutral class did not exist any apparent class-level information recognized by humans, but CSS-LM can distinguish these instances in the sentence-level; however, CSS-LM takes the instance, \textit{$...$ there sister restaurant arties cafe}, as the positive, which can be easily classified to the neutral class by humans with sentence-level meaning. To consider the sentence-level information during retrieving by contrastive learning is a future work we can study.

\section{Conclusion and Future Work}

In this work, we introduce the CSS-LM framework to improve the fine-tuning phase of PLMs via contrastive semi-supervised learning. The experimental results on three typical text classification tasks show that CSS-LM could effectively capture crucial semantic features for downstream tasks with limited supervised data and achieve better performances than the conventional, supervised contrastive fine-tuning strategies. In future, we will explore the following promising directions: 

(1) The CSS-LM framework makes an initial attempt to better fine-tune PLMs with limited supervised data of downstream tasks in text classification. Extending it to other NLP tasks, e.g., text generation, name entity recognition, and question answering, is a valuable direction. 

(2) From our experimental results, we find that better methods to consider domain-level semantic remain a further exploration. 
Finding out a better strategy to retrieve and denoise instances is also a fascinating problem. 

(3) CSS-LM is devoted to leveraging unannotated data from the open-domain corpora to capture crucial semantic features, which benefit the downstream tasks, instead of considering invariant features between different tasks or domains to perform transfer learning. How to better leverage these invariant features is a direction we can explore.


\ifCLASSOPTIONcompsoc
  \section*{Acknowledgments}
\else
  \section*{Acknowledgment}
\fi
This work is supported by the National Key Research and Development Program of China (No. 2020AAA0106501) and the National Natural Science Foundation of China (NSFC No. 61772302).

\ifCLASSOPTIONcaptionsoff
  \newpage
\fi




\bibliographystyle{IEEEtran}
\bibliography{TKDE}

\end{document}